\pdfoutput=1

\documentclass[11pt]{article}

\usepackage[final]{latex/acl}

\usepackage{enumitem}
\usepackage{times}
\usepackage{latexsym}

\usepackage{listings}
\lstdefinestyle{base}{
  emptylines=1,
  breaklines=true,
  basicstyle=\ttfamily\color{black},
  moredelim=**[is][\color{red}]{@}{@},
  moredelim=**[is][\color{green}]{*}{*}
}
\usepackage[T1]{fontenc}

\usepackage[utf8]{inputenc}

\usepackage{microtype, ifsym, marvosym}

\usepackage{microtype}

\usepackage{inconsolata}

\usepackage{graphicx}
\usepackage{hyperref}
\usepackage{url}
\usepackage{caption}
\usepackage{pdfpages}
\usepackage{float}
\usepackage{multirow}
\usepackage{multicol}
\usepackage{makecell}
\usepackage{array}
\usepackage{booktabs}
\usepackage{caption}
\usepackage{amsmath}
\usepackage{textcomp}
\usepackage[accsupp]{axessibility}

\usepackage{array}

\usepackage{multirow}
\usepackage[usestackEOL]{stackengine}

\usepackage{xspace}

\usepackage{pifont}%
\newcommand{\cmark}{\textcolor[rgb]{0.13, 0.55, 0.13}{\ding{51}}}%
\newcommand{\xmark}{\textcolor[rgb]{0.5, 0, 0}{\ding{55}}}%

\newcommand{\ie}{\textit{i.e.}\xspace}

\newcommand{\etc}{\textit{etc.}\xspace}

\title{Evaluation Agent: Efficient and Promptable Evaluation Framework for Visual Generative Models}

\author{%
  Fan Zhang$^{1*}$, Shulin Tian$^{2*}$, Ziqi Huang$^{2\ddag*}$,  Yu Qiao\textsuperscript{1\Letter}, Ziwei Liu\textsuperscript{2\Letter} \\
   \\
   $^{1}$Shanghai Artificial Intelligence Laboratory, $^{2}$S-Lab, Nanyang Technological University \\
   $^*$Equal Contributions. $^{\ddag}$ Project Lead. \textsuperscript{\Letter}Corresponding Authors.\\
  \small\texttt{zhangfan199903@163.com, \{shulin002, ziqi002\}@ntu.edu.sg}\\
  \\
  [-10pt]
  \tt\normalsize\url{https://vchitect.github.io/Evaluation-Agent-project/}
}

\begin{document}
\maketitle


\footnotetext{\hspace{-15pt}\href{https://github.com/Vchitect/Evaluation-Agent} {Code} is available}\makeatother

\begin{abstract}

Recent advancements in visual generative models have enabled high-quality image and video generation, opening diverse applications. However, evaluating these models often demands sampling hundreds or thousands of images or videos, making the process computationally expensive, especially for diffusion-based models with inherently slow sampling. Moreover, existing evaluation methods rely on rigid pipelines that overlook specific user needs and provide numerical results without clear explanations. In contrast, humans can quickly form impressions of a model's capabilities by observing only a few samples. To mimic this, we propose the \textbf{Evaluation Agent} framework, which employs human-like strategies for efficient, dynamic, multi-round evaluations using only a few samples per round, while offering detailed, user-tailored analyses. It offers four key advantages: \textit{1)} \textbf{efficiency}, \textit{2)} \textbf{promptable evaluation} tailored to diverse user needs, \textit{3)} \textbf{explainability} beyond single numerical scores, and \textit{4)} \textbf{scalability} across various models and tools. Experiments show that Evaluation Agent reduces evaluation time to 10\% of traditional methods while delivering comparable results. The Evaluation Agent framework is fully open-sourced to advance research in visual generative models and their efficient evaluation.

\begin{figure}[t!]
    \centering
    \includegraphics[width=\linewidth]{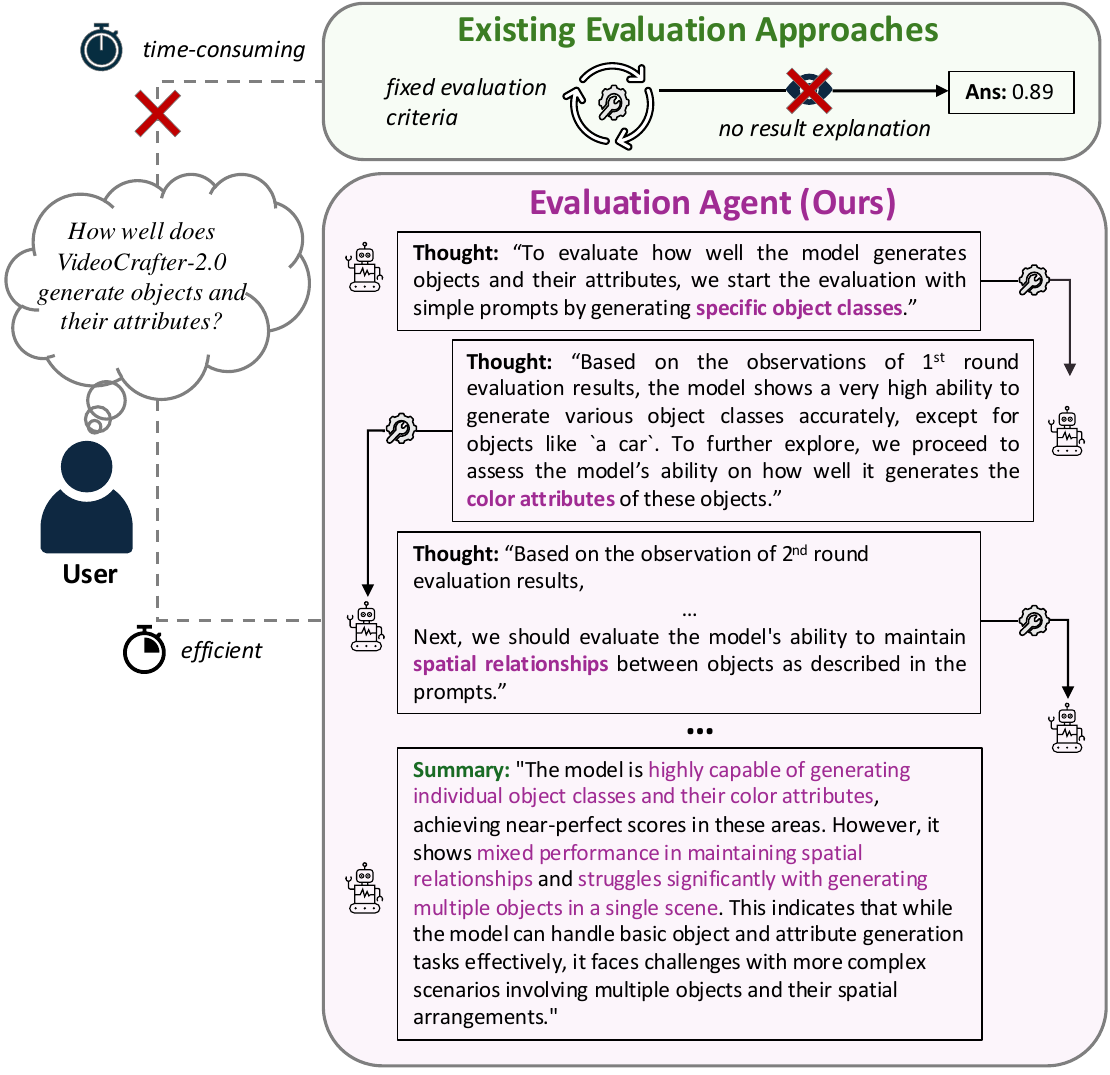}
    \caption{\textbf{An Example of Evaluation Agent.} Existing evaluation methods typically assess visual generative models by extensively sampling from a fixed benchmark. In contrast, our Evaluation Agent framework requires only a small number of sampled images or videos, tailored to the user's specific evaluation request. Additionally, it goes beyond providing a simple numerical score by offering detailed explanations to the evaluation conclusions.}
    \label{fig:use_case}
    \vspace{-20pt}
\end{figure}

\end{abstract}

\section{Introduction}
\label{intro}

Visual generative models have made significant progress in recent years, particularly driven by the advancement of diffusion models~\cite{ho2020denoising} and the availability of internet-scale datasets~\cite{Bain21, chen2024panda70m, xue2019video}. These advancements enable the generation of high-quality images and videos, opening up a wide range of applications in content creation, design inspiration, and beyond.

With the advancement of visual generative models, effective evaluation is crucial for understanding their strengths, limitations, and areas for improvement. Existing evaluation frameworks, such as VBench~\cite{huang2023vbench, huang2024vbench++}, EvalCrafter~\cite{liu2023evalcrafter}, and T2I-CompBench~\cite{huang2023t2icompbench}, assess models across multiple dimensions using specific prompts and tailored metrics to ensure comprehensive performance analysis. However, these approaches often demand generating numerous samples, resulting in long evaluation time and high computational costs, particularly for diffusion-based models where sampling is inherently slow due to iterative sampling. Furthermore, these evaluation frameworks are constrained by rigid evaluation pipelines and predefined dimensions, making them less adaptable to open-ended inputs or diverse user needs. Additionally, these methods often produce single numerical scores as outcomes, requiring users to invest additional effort to extract meaningful insights.

In contrast, human evaluators can quickly gain a general understanding of a model’s performance by interactively testing a few prompts, forming a sufficient impression without taking too much time. This type of evaluation has several unique advantages for assessing visual generative models. First, it is \textbf{fast}, requiring only a small number of samples to assess overall performance. Second, it is \textbf{flexible}, allowing intuitive evaluation of various aspects, such as realism, creativity, prompt adherence, or other user-defined criteria. Third, it is \textbf{dynamic}, enabling deeper, hierarchical evaluations through continuous adjustments in exploration.

To leverage the strengths of human-like evaluations, we introduce the \textbf{Evaluation Agent}, a paradigm that mimics human strategies for assessing visual generative models. The Evaluation Agent offers four key features: \textbf{\textit{1) Efficiency}:} It dynamically adjusts its evaluation pathway based on intermediate results, uncovering subtle model behaviors and limitations while avoiding redundant test cases for efficient evaluation. \textbf{\textit{2) Promptable Evaluation}:} Unlike existing benchmarks with fixed prompts and evaluation metrics, it accepts open-ended user input, allowing for flexible and customized assessments tailored to specific user needs. \textbf{\textit{3) Detailed and Interpretable Results}:} It provides interpretable, detailed insights beyond single numerical scores, making results accessible to both experts and non-experts. \textbf{\textit{4) Scalability}:} The framework supports seamless integration of new metrics and evaluation tools, ensuring adaptability and growth.

The Evaluation Agent begins by accepting open-ended user input, specifying what to evaluate and which model(s) to assess. Based on this input, it identifies initial evaluation aspects and leverages appropriate tools to conduct the assessment. It then observes the intermediate results and dynamically refines the direction of further exploration. In the end, it generates a detailed natural language response summarizing the evaluation results, providing a comprehensive analysis of the evaluation process and a clear summary of the model’s capabilities as specified in the user input. 
The Evaluation Agent can also automate various applications, including: \textit{1) Model Comparison}: Allowing users to compare models based on specific criteria to determine which performs better in a given aspect. \textit{2) Model Recommendation}: Suggesting the most suitable model for the user’s needs by evaluating models against personalized criteria.

\begin{table*}[t]
\centering
\resizebox{\textwidth}{!}{%
\begin{tabular}{c|c|c|c|c|c|c|c}
\toprule
\textbf{Benchmark} & \textbf{\makecell{Analysis}} & \textbf{\Centerstack{Customized \\Queries}} & \textbf{\Centerstack{Supported \\Models}} & \textbf{\Centerstack{\# Required\\Samples}} & \textbf{\Centerstack{Open Evaluation\\Request Support}} & \textbf{\Centerstack{Dynamic \\Evaluation}} & \textbf{\Centerstack{Open \\Tool-Use}} \\
\midrule
FID / FVD~\cite{unterthiner2018towards, heusel2017gans} & \xmark & \xmark & T2I / T2V & 2,048 & \xmark~(Fixed-Form) & \xmark & \xmark\\ 
T2I-CompBench~\cite{huang2023t2icompbench} & \xmark & \xmark & T2I & 18,000 & \xmark~(Pre-Defined) & \xmark & \xmark\\
VBench~\cite{huang2023vbench} & \xmark & \xmark & T2V & 4,730 & \xmark~(Pre-Defined) & \xmark & \xmark\\
\midrule
\textbf{Evaluation Agent (Ours)} & \cmark & \cmark & T2I \& T2V & 4\~00 & \cmark~(Open-Ended) & \cmark & \cmark\\
\bottomrule
\end{tabular}%
}
\caption{\textbf{Comparison of the Evaluation Agent Framework with Traditional T2I and T2V Benchmarks.} The Evaluation Agent framework supports customized user queries in natural language and works with both T2I and T2V models. Unlike traditional benchmarks, it dynamically updates the evaluation process using multiple tools, providing comprehensive and explainable results with detailed textual analysis.}
\label{tab:related_work_comp}
\vspace{-15pt}
\end{table*}

We demonstrate the versatility of the Evaluation Agent through experiments on diverse scenarios, including the evaluation of image and video generation models. The results indicate that it delivers performance comparable to full benchmark pipelines while significantly reducing evaluation time. Furthermore, we create an open-ended user query dataset to showcase the Evaluation Agent’s flexibility, depth, and accuracy in addressing open-ended queries.

We summarize our contributions as follows:
\begin{itemize}[noitemsep, topsep=0pt] \item We propose the \textbf{Evaluation Agent}, a human-like evaluation framework that overcomes the limitations of existing methods in flexibility and efficiency. Our approach will be fully open-sourced. \item We validated our approach on several widely adopted benchmarks, demonstrating that it achieves evaluation accuracy comparable to standard benchmarks while reducing evaluation time by over 90\%. We also built an open-ended user query dataset to demonstrate our method’s flexibility, depth, and accuracy in handling open-ended evaluation queries. \item Through a comprehensive analysis of how standard benchmarks, human evaluators, and our Evaluation Agent perform evaluations, we provide in-depth insights that serve as important cornerstones for future research in evaluating visual generative models. \end{itemize}

\section{Related Work}
\label{sec:related}

\subsection{Visual Generation and Evaluation}

Visual generative models have gained significant attention in recent years. However, unlike perception tasks, which have clear evaluation metrics such as accuracy, evaluating visual generative tasks is more challenging due to the absence of a definitive “ground truth” or single correct answer. Metrics such as FID~\cite{heusel2017gans} and FVD~\cite{unterthiner2018towards} are commonly used to measure the distance between generated samples and reference datasets. Recent benchmarks~\cite{huang2023t2icompbench, huang2023vbench, zheng2025vbench2, huang2024vbench++, liu2023evalcrafter, lee2023holistic, sun2024t2v}  provide multi-dimensional evaluations tailored to specific model capabilities. However, whether relying on metrics like FID~\cite{heusel2017gans} and FVD~\cite{unterthiner2018towards} or benchmarks like VBench~\cite{huang2023vbench}, these evaluations are often time-intensive and limited in scope.

\subsection{LLM as a Judge}
Recently, the development of understanding and reasoning capabilities in Large Language Models (LLMs) demonstrates a significant advantage, enabling them to serve as powerful evaluators~\cite{Jain_2023, chiang2023largelanguagemodelsalternative, fu2023gptscoreevaluatedesire, zheng2023judging}. For instance, CoEval~\cite{li2023collaborativeevaluationexploringsynergy} introduces a two-stage evaluation framework for open-ended natural language generation (NLG) tasks, offering a scalable and cost-efficient alternative to human evaluations. \citet{pan2024autonomousevaluationrefinementdigital} demonstrates how LLM-based evaluations enhance downstream tasks for digital agents. These studies showcase the ability of LLMs to reason, explain, and comprehend evaluation processes. Despite these advancements, prior work primarily focuses on improving general reasoning or minimizing hallucination, leaving the use of LLMs for evaluating visual generative models largely unexplored. Furthermore, though LLMs demonstrate considerable proficiency in zero-shot reasoning and planning, devising effective strategies for domain-specific problems remains challenging~\cite{wang2024survey}, complicating their role as evaluators for visual generative tasks.

\subsection{Agent in Planning \& Reasoning}

Agents and agentic systems are gaining attention for their ability to automate complex tasks and design customized trajectories based on user queries. They have been explored across various domains, including web, mobile, desktop, and operating systems (OS) ~\cite{zhou2023webarena, xie2024osworld, wang2024mobile, kapoor2024omniact, zhang2024mmina}, showing effectiveness in improving long-horizon task completion. For example, Chain-of-Thought (CoT)~\cite{wei2022chain} and Zero-shot-CoT~\cite{kojima2022large} use prompting techniques to enable step-by-step reasoning. Similarly, ReAct~\cite{yao2022react} introduces a general paradigm for agent prompting by integrating reasoning traces with task-specific actions through interleaved triplets of “thought-action-observation,” thereby incorporating environmental feedback. Despite the broad applicability of agents in task automation, their potential to automate the evaluation process of visual generative models remains largely unexplored.
\section{Methods}
\label{methods}

\begin{figure*}[htbp!]
\centering
\includegraphics[width=\textwidth]{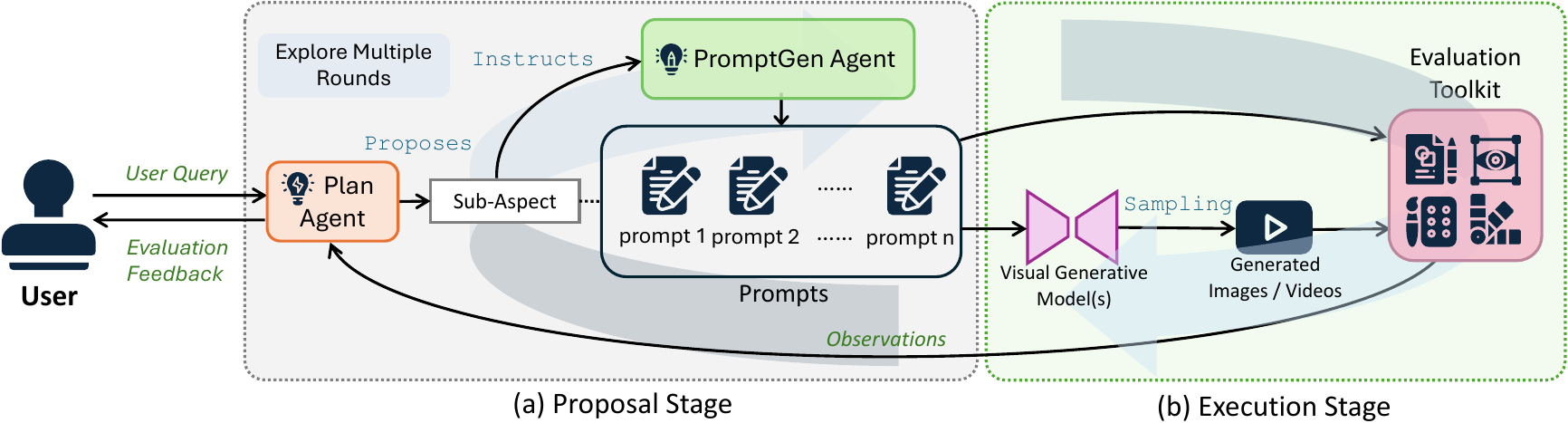}
\caption{\textbf{Overview of Evaluation Agent Framework.} This framework leverages LLM-powered agents for efficient and flexible visual model assessments. As shown, it consists of two stages: (a) the Proposal Stage, where user queries are decomposed into sub-aspects, and prompts are generated, and (b) the Execution Stage, where visual content is generated and evaluated using an Evaluation Toolkit. The two stages interact iteratively to dynamically assess models based on user queries.}
\label{fig:pipeline}
\vspace{-15pt}
\end{figure*}

\subsection{Preliminaries: Evaluation of Visual Generative Models}

\noindent\textbf{Evaluation Benchmark.} $C=\{c_{j} | j \in \{1, 2, 3, ...., N\}\}$, where the conditions (\ie, test cases) $C$ is a set of text prompts, input images, class labels, or conditionals in other formats. In the case of unconditional generation, $C$ is empty or consists of random seeds for unconditional generation. In existing evaluation approaches, $C$ is pre-defined and usually contains at least hundreds or thousands of items, which require intensive computation and time cost for sampling. In our Evaluation Agent framework, the test case set $C$ is dynamically determined during the evaluation process, and usually only contains a few cases towards the end.

\noindent\textbf{Sampling.} $v_{j} = G(c_{j})$
where $G$ is the visual generative model, which generates the visual output $v_j$ (\ie, images or videos) given an optional condition $c_j$. $V = G(C)$
where $V=\{v_j | j \in \{1, 2, 3, ...., N\}\}$ represents the set of generated visuals for the entire condition set $C$.

\noindent\textbf{Evaluation Pipeline.} Existing evaluation methods usually follow a fixed pipeline to evaluate all the images or videos $V$ sampled from the pre-defined benchmark $C$.
\begin{equation}
    y_j = e_k(v_j, c_j)
\end{equation}
where $e_k \in E$ is an evaluation function of some aspects such as aesthetic, compositionality \etc. Given the generated visual $v_j$,  and the optional condition $c_j$, it produces the evaluation result $y_j$.

Some reference/statistics-based evaluation frameworks like FID and FVD also use reference datasets $V_r$ for calculating the results.
\begin{equation}
    Y = E(V, V_r, C)
\end{equation}
In existing evaluation approaches, $E$ is a pre-defined set of evaluation dimensions, or a single evaluation metric, which limits the possible evaluation aspects from the beginning, and requires assessing all the aspects even if some are not needed in some cases. In our Evaluation Agent framework, the evaluation tool $e_k$ is dynamically determined during the evaluation process.

\begin{table*}[t]
\captionof{table}{\textbf{Evaluation Results Comparison with VBench~\cite{huang2023vbench}}. We evaluated 15 specific ability dimensions in VBench using our Evaluation Agent and compared its results against VBench in terms of conclusion accuracy. The numerical results show the percentages of the Evaluation Agent’s correct predictions falling either within the exact range (left) or within an error margin of one range (right) across ten trials.}
    \centering
    \small
    \setlength{\tabcolsep}{2pt}
    \resizebox{\textwidth}{!}{%
        \begin{tabular}{>{\centering\arraybackslash}p{4.5cm}|
                            >{\centering\arraybackslash}p{2cm}|
                            >{\centering\arraybackslash}p{2cm}|
                            >{\centering\arraybackslash}p{2cm}|
                            >{\centering\arraybackslash}p{2cm}|
                            >{\centering\arraybackslash}p{2cm}|
                            >{\centering\arraybackslash}p{2cm}|
                            >{\centering\arraybackslash}p{2cm}}
        \Xhline{1pt}
        \textbf{Models} & \textbf{\Centerstack{Subject\\Consistency}} & \textbf{\Centerstack{Background\\Consistency}} & 
        \textbf{\Centerstack{Motion\\Smoothness}} & 
        \textbf{\Centerstack{Dynamic\\Degree}} & 
        \textbf{\Centerstack{Aesthetic\\Quality}} & 
        \textbf{\Centerstack{Imaging\\Quality}} & 
        \textbf{\Centerstack{Object\\Class}} \\
        \Xhline{1pt}
        Latte-1~\cite{ma2024latte} & 50{\%} / 80{\%} & 0{\%} / 30{\%} & 40{\%} / 70{\%} & 30{\%} / 70{\%} & 60{\%} / 100{\%} & 70{\%} / 100{\%} & 40{\%} / 50{\%} \\ 
        ModelScope~\cite{wang2023modelscope}   & 80{\%} / 80{\%} & 80{\%} / 90{\%} & 60{\%} / 80{\%} & 60{\%} / 100{\%} & 60{\%} / 100{\%} & 100{\%} / 100{\%} & 0{\%} / 50{\%} \\ 
        VideoCrafter-0.9~\cite{he2022lvdm}  & 100{\%} / 100{\%} & 80{\%} / 100{\%} & 70{\%} / 100{\%} & 80{\%} / 100{\%} & 90{\%} / 100{\%} & 20{\%} / 100{\%} & 20{\%} / 60{\%} \\ 
        VideoCrafter-2~\cite{chen2024videocrafter2}  & 10{\%} / 100{\%} & 60{\%} / 100{\%} & 30{\%} / 90{\%} & 30{\%} / 80{\%} & 80{\%} / 100{\%} & 50{\%} / 100{\%} & 70{\%} / 100{\%} \\ 
        \Xhline{1pt}    
        \end{tabular}
    }

    \vspace{5pt}

    \setlength{\tabcolsep}{2pt}
    \resizebox{\textwidth}{!}{%
    \begin{tabular}{>{\centering\arraybackslash}p{4.5cm}|
                            >{\centering\arraybackslash}p{2cm}|
                            >{\centering\arraybackslash}p{2cm}|
                            >{\centering\arraybackslash}p{2cm}|
                            >{\centering\arraybackslash}p{2cm}|
                            >{\centering\arraybackslash}p{2cm}|
                            >{\centering\arraybackslash}p{2cm}|
                            >{\centering\arraybackslash}p{2cm}}
    \Xhline{1pt}
    \textbf{\Centerstack{Multiple\\Objects}} &
    \textbf{\Centerstack{Human\\Action}} & 
    \textbf{\Centerstack{Color}} & 
    \textbf{\Centerstack{Spatial\\Relationship}} & 
    \textbf{\Centerstack{Scene}} & 
    \textbf{\Centerstack{Appearance\\Style}} & 
    \textbf{\Centerstack{Temporal\\Style}} & 
    \textbf{\Centerstack{Overall\\Consistency}} \\
    \Xhline{1pt}
    40{\%} / 100{\%} & 10{\%} / 10{\%} & 30{\%} / 70{\%} & 10{\%} / 80{\%} & 20{\%} / 40{\%} & 70{\%} / 90{\%} & 40{\%} / 100{\%} & 70{\%} / 100{\%} \\
    50{\%} / 100{\%} & 10{\%} / 40{\%} & 0{\%} / 20{\%} & 10{\%} / 30{\%} & 20{\%} / 100{\%} & 90{\%} / 100{\%} & 50{\%} / 90{\%} & 20{\%} / 100{\%} \\

    80{\%} / 100{\%} & 10{\%} / 30{\%} & 10{\%} / 40{\%} & 20{\%} / 100{\%} & 30{\%} / 100{\%} & 60{\%} / 100{\%} & 80{\%} / 100{\%} & 0{\%} / 80{\%} \\
    20{\%} / 60{\%} & 10{\%} / 90{\%} & 90{\%} / 100{\%} & 0{\%} / 70{\%} & 0{\%} / 10{\%} & 80{\%} / 100{\%} & 80{\%} / 100{\%} & 60{\%} / 100{\%} \\
    \Xhline{1pt}
    \end{tabular}
    }
    \label{tab:main_results_t2v}
    \vspace{-5pt}
\end{table*}

\begin{table*}[htbp]
    \caption{\textbf{Evaluation Results Comparison with T2I-CompBench~\cite{huang2023t2icompbench}}. We evaluated four ability dimensions in T2I-CompBench using our Evaluation Agent and compared its results with those of T2I-CompBench in terms of conclusion accuracy. The numerical results show the percentages of the Evaluation Agent’s correct predictions falling either within the exact range (left) or within an error margin of one range (right) across ten trials.}    \centering
    \small
    \setlength{\tabcolsep}{3pt}
    \begin{tabular}{c|c|c|c|c}
    \Xhline{1pt}
    \textbf{Models} & \textbf{\Centerstack{Color\\Binding}} &     \textbf{\Centerstack{Shape\\Binding}} &
    \textbf{\Centerstack{Texture\\Binding}} &     \textbf{\Centerstack{Non-Spatial\\Relationships}} \\ \Xhline{1pt}
    SD1.4~\cite{rombach2022high}        & 50{\%} / 100{\%} & 100{\%} / 100{\%} & 0{\%} / 100{\%} & 50{\%} / 100{\%} \\ 
    SD2.1 ~\cite{rombach2022high}  & 100{\%} / 100{\%} & 60{\%} / 100{\%} & 80{\%} / 100{\%} & 60{\%} / 100{\%} \\ 
    SDXL ~\cite{podell2023sdxl}  & 100{\%} / 100{\%} & 20{\%} / 100{\%} & 80{\%} / 100{\%} & 60{\%} / 100{\%} \\ 
    SD3.0 ~\cite{esser2024scaling}  & 20{\%} / 90{\%} & 0{\%} / 90{\%} & 0{\%} / 70{\%} & 80{\%} / 90{\%} \\ 
    \Xhline{1pt}
    \end{tabular}
    \vspace{-10pt}
    \label{tab:main_results_t2i}
\end{table*}

\subsection{The Evaluation Agent Framework}

Our Evaluation Agent framework is powered by LLM-based agents, leveraging their advanced planning capabilities to simulate human-like behaviors for efficient and flexible visual model assessments. As illustrated in Figure~\ref{fig:pipeline}, the framework operates in two stages: the proposal stage and the execution stage. By iteratively interacting and looping between these stages, the framework dynamically evaluates models based on user queries.

\subsubsection{Proposal Stage}
\label{sssec:proposal_stage}

The Proposal Stage consists of two agents: the Plan Agent and the PromptGen Agent. The Plan Agent is responsible for planning, observing, and summarizing the evaluation process based on the user’s query, while the PromptGen Agent focuses specifically on the design aspects.

\noindent\textbf{Plan Agent.} We design the Plan Agent to simulate human behavior during the evaluation process, including planning and adjusting the evaluation direction, observing intermediate results, and summarizing the final outcomes. As the core component of the framework, the Plan Agent not only interacts with the user but also drives the entire evaluation process. Specifically, upon receiving a user query, the Plan Agent identifies an initial sub-aspect for evaluation and iteratively refines it based on feedback from intermediate results. This process continues until sufficient information is collected, after which the agent provides a detailed analysis and summary. Additional details can be found in Appendix~\ref{appendix a}.

\noindent\textbf{PromptGen Agent.} The PromptGen Agent mimics human behavior in designing prompts for visual generative models based on the plan developed during the evaluation process. Specifically, it generates tailored prompts for each sub-aspect identified by the Plan Agent, enabling focused content generation and evaluation. Additionally, the PromptGen Agent can reference and utilize well-established prompts from existing benchmarks. Further details are provided in Appendix~\ref{appendix a}.

\subsubsection{Execution Stage}
\label{sssec:execution_stage}

The Execution Stage is responsible for sampling and evaluating the model using the appropriate tools, as specified in the Proposal Stage, and for returning the final evaluation results.

\noindent\textbf{Visual Generative Models.} This component takes prompts designed by the PromptGen Agent as input and generates corresponding visual content, which is then used for subsequent evaluation.

\noindent\textbf{Evaluation Toolkit.} The Evaluation Toolkit consists of a set of elementary evaluation tools for visual generative models. This module is open and extensible, allowing for continuous expansion. We have integrated several existing evaluation tools from well-known benchmarks for different modalities of visual generation models. To support the evaluation of open-ended user queries, we have introduced a paradigm based on vision-language models (VLMs), leveraging the Visual Question Answering (VQA) format to enable flexible assessments across various aspects of the models. Upon receiving the generated samples from visual generative models along with the corresponding prompts, the module utilizes the appropriate tools to evaluate each sample. All evaluation results are then compiled and returned to the Plan Agent for further proposals or summarization. For detailed information, please refer to the Appendix~\ref{appendix a}.

\subsubsection{Overall Pipeline}
\label{sssec:overall_pipeline}

The Evaluation Agent’s process is dynamic and multi-round, with each round comprising a proposal stage and an execution stage. By interacting and looping through these stages, we achieve dynamic evaluation, where the evaluation process adapts based on intermediate observations and initial user query. This dynamic approach allows the Evaluation Agent to refine its focus iteratively, adjusting its exploration direction and prompt design based on an evolving understanding of the model’s capabilities. Consequently, the evaluation process becomes more efficient and targeted, systematically identifying the strengths and limitations of generative models.

\section{Experiments}
\label{sec:experiments}

\begin{table*}[htbp]
    \centering
    \setlength\tabcolsep{3pt}
    \small
    \caption{\textbf{Time Cost Comparison across Models for VBench Dimensions.} This table compares the evaluation time of four different models using the original VBench pipelines versus the Evaluation Agent. The Evaluation Agent significantly reduces the overall evaluation time.}
    \resizebox{0.99\linewidth}{!}{
    \begin{tabular}{c||c|c||c}
        \Xhline{1pt}
        \textbf{Models}   & \textbf{\Centerstack{VBench (Total Cost) $ \downarrow $}} & \textbf{\Centerstack{VBench (Avg. Cost per Dimension) $ \downarrow $}} & \textbf{\Centerstack{Evaluation Agent (Ours) $ \downarrow $}} \\ \Xhline{1pt}
        Latte-1~\cite{ma2024latte} & 2557 min, 4355 samples & 170 min, 290 samples & 15 min, 25 samples  \\ 
        ModelScope~\cite{wang2023modelscope} & 1160 min, 4355 samples & 77 min, 290 samples & 6 min, 23 samples  \\ 
        VideoCrafter-0.9~\cite{he2022lvdm}  & 1459 min, 4355 samples & 97 min, 290 samples & 9 min, 24 samples  \\ 
        VideoCrafter-2~\cite{chen2024videocrafter2}  & 4261 min, 4355 samples & 284 min, 290 samples & 24 min, 23 samples  \\  \hline
    \end{tabular}
    }
    \vspace{-5pt}
    \label{tab:t2v_time}
\end{table*}

\begin{table*}[t]
    \centering
    \setlength\tabcolsep{3pt}
    \small
    \caption{\textbf{Time Cost Comparison across Models for T2I-CompBench Dimensions.} This table compares the evaluation costs for assessing four models across T2I-CompBench dimensions using both the original T2I-CompBench pipelines and our Evaluation Agent. The Evaluation Agent achieves a substantial reduction in evaluation time compared to the traditional pipelines.
    }
    \resizebox{0.99\linewidth}{!}{
    \begin{tabular}{c||c|c||c}
        \Xhline{1pt}
        \textbf{Models}   & \textbf{\Centerstack{T2I-Comp (Total Cost) $ \downarrow $}} & \textbf{\Centerstack{T2I-Comp (Avg. Cost per Dimension) $ \downarrow $}} & \textbf{\Centerstack{Evaluation Agent (Ours) $ \downarrow $}} \\ \Xhline{1pt}
        SD1.4~\cite{rombach2022high} & 563 min, 12000 samples & 141 min, 3000 samples & 5 min, 26 samples  \\ 
        SD2.1~\cite{rombach2022high} & 782 min, 12000 samples & 196 min, 3000 samples & 6 min, 26 samples  \\ 
        SDXL~\cite{podell2023sdxl}  & 1543 min, 12000 samples & 386 min, 3000 samples & 8 min, 26 samples  \\ 
        SD3.0~\cite{esser2024scaling}  & 1410 min, 12000 samples & 353 min, 3000 samples & 7 min, 25 samples  \\  \hline
    \end{tabular}
    }
    \vspace{-10pt}
    \label{tab:t2i_time}
\end{table*}

We first validate the efficiency of our Evaluation Agent on established benchmarks for visual generative models and then demonstrate the flexibility, depth, and accuracy of our approach in handling open-ended user queries on our self-constructed dataset.

\subsection{Experiments on Existing Benchmarks}
\label{ssec:closed-domain}

We validate the effectiveness of our framework on both the Text-to-Video (T2V) and Text-to-Image (T2I) tasks. For detailed settings and implementations, please refer to Appendix~\ref{appendix b}.

\subsubsection{Experimental Setup}

\noindent\textbf{Visual Generative Models.}
For the T2V task, we select four open-source models: VideoCrafter-0.9~\cite{he2022lvdm}, VideoCrafter-2~\cite{chen2024videocrafter2}, Latte-1~\cite{ma2024latte}, and ModelScope~\cite{wang2023modelscope}. Similarly, for the T2I task, we choose four well-known open-source models: SD(Stable Diffusion)1.4~\cite{rombach2022high}, SD2.1~\cite{rombach2022high}, SDXL~\cite{podell2023sdxl}, and SD3.0~\cite{esser2024scaling}. We validate the efficiency and accuracy of our method using these models within the evaluation frameworks for both T2I and T2V tasks.

\noindent\textbf{Visual Generation Benchmarks.}
For the T2V and T2I tasks, we adopt well-established and comprehensive evaluation frameworks tailored to their respective domains. Specifically, for the T2V task, we use VBench~\cite{huang2023vbench}, a comprehensive and fine-grained evaluation framework for video generation. For the T2I task, we select T2I-CompBench~\cite{huang2023t2icompbench}, which assesses the compositional generation capabilities of T2I models across multiple dimensions. We validate the effectiveness of our method using these evaluation frameworks.

\noindent\textbf{Comparison Setup.}
To ensure comparability and fairness in our experiments, we restrict the evaluation tools and prompts available to the Evaluation Agent. For T2V tasks, we used only the metrics and prompt lists from VBench, and for T2I tasks, only those from T2I-CompBench. For comparison, we categorize performance into five levels based on the performance density distribution. For more details, see Appendix~\ref{appendix b}.

\subsubsection{Results Analysis}
\label{sssec:t2v_results}

\noindent\textbf{Validation on VBench.} To highlight the efficiency of our proposed methods, we compare the time consumption and sample counts between VBench and our approach. As shown in Table~\ref{tab:t2v_time}, our method reduces evaluation time by over 10X. Additionally, Table~\ref{tab:main_results_t2v} confirms the consistency of our evaluation results with VBench across various dimensions. 
The quantitative results show that the Evaluation Agent achieves high prediction accuracy across most dimensions, demonstrating that our approach maintains accuracy while significantly reducing evaluation time.

\begin{figure}
    \vspace{2pt}
    \centering
    \includegraphics[width=0.85\linewidth]{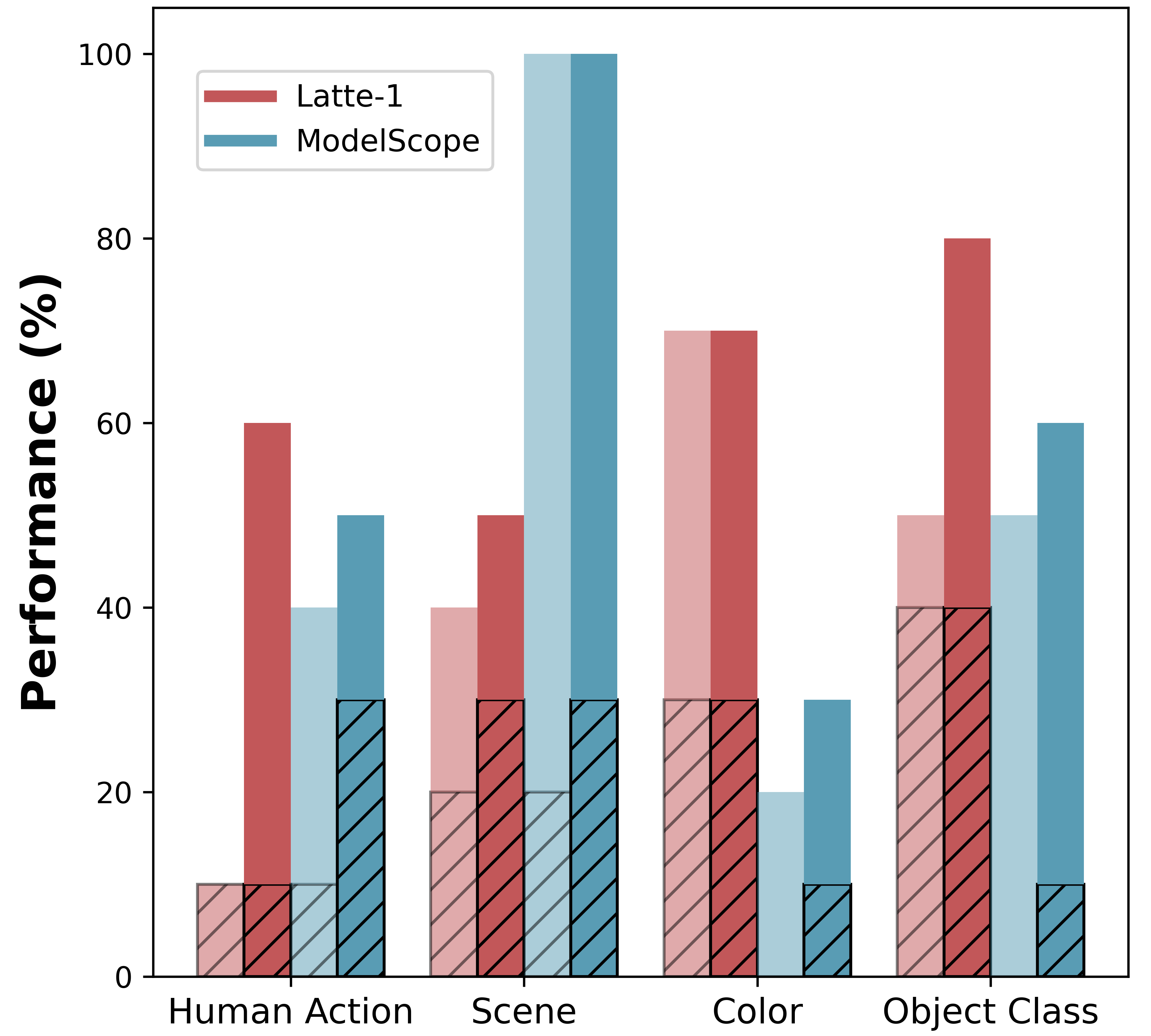} 
    \caption{\textbf{Validation on VBench Percentage Dimensions.} We conducted additional validation experiments on VBench by increasing the number of prompts in each evaluation. For each model and dimension, lighter bars represent results with the original settings, darker bars with increased sample size. Hatched portions indicate predictions within the exact range, and solid portions within an error margin of one range. Specific numerical results are provided in Table~\ref{tab:t2v_ablation}}
    \label{fig:t2v_percent_bar}
    \vspace{-10pt}
\end{figure}

\noindent\textbf{Validation on T2I-CompBench.}
\label{sssec:t2i_results}
We evaluate the Evaluation Agent’s performance on T2I tasks, as shown in Table~\ref{tab:t2i_time}. Instead of requiring thousands of samples and several hours, the Evaluation Agent completes evaluations with just about 26 samples in five to eight minutes per dimension. Table~\ref{tab:main_results_t2i} compares the evaluation results, showcasing high accuracy within an error margin of one range.

\begin{table*}[htp]
    \centering
    \setlength\tabcolsep{3pt}
    \small
    \caption{\textbf{Validation on VBench Percentage Dimensions.} The numerical results show the percentages of the Evaluation Agent’s correct predictions falling either within the exact range (left) or within an error margin of one range (right) across ten trials.}
    \begin{tabular}{c|c|c|c|c}
        \Xhline{1pt}
        \textbf{Models}   & \textbf{\Centerstack{Human\\Action}} & \textbf{\Centerstack{Scene}} & 
        \textbf{\Centerstack{Color}} & \textbf{\Centerstack{Object\\Class}} \\ \Xhline{1pt}
        Latte-1 (default)~\cite{ma2024latte} & 10{\%} / 10{\%} & 20{\%} / 40{\%} & 30{\%} / 70{\%} & 40{\%} / 50{\%}   \\ 
        Latte-1 (30 prompts)~\cite{ma2024latte} & 10{\%} / 60{\%} & 30{\%} / 50{\%} & 30{\%} / 70{\%} & 40{\%} / 80{\%}   \\ \hline
        ModelScope (default)~\cite{wang2023modelscope}  & 10{\%} / 40{\%} & 20{\%} / 100{\%} & 0{\%} / 20{\%} & 0{\%} / 50{\%} \\ 
        ModelScope (30 prompts)~\cite{wang2023modelscope}  & 30{\%} / 50{\%} & 30{\%} / 100{\%} & 10{\%} / 30{\%} & 10{\%} / 60{\%} \\ \hline
    \end{tabular}
    \vspace{-10pt}
    \label{tab:t2v_ablation}
\end{table*}

\noindent\textbf{Analysis of Percentage-Based Dimensions.}
\label{sssec:percentage_results}
In the validation experiments on VBench, we observe lower evaluation results in dimensions like ``Human Action'', ``Scene'', ``Color'', and ``Object Class''. Further analysis reveals that these dimensions rely on statistical evaluations, where sample-level results are binary (0 or 1), making them highly sensitive to sample size. Since the Evaluation Agent limits samples per round to three to nine, we test this sensitivity by increasing the sample size to 30 using the ModelScope~\cite{wang2023modelscope} and Latte-1~\cite{ma2024latte} models. The results, shown in Figure~\ref{fig:t2v_percent_bar}, indicate that as the sample size increases, the performance of both models in each dimension steadily improves.

\noindent\textbf{More Results.}
The experiments described above were conducted using the GPT model as the backbone. To further validate the effectiveness and versatility of our framework, we extended this set of experiments to other models, including Claude and Gemini. Detailed results and analyses can be found in Appendix~\ref{appendix e1}.

\begin{figure}
    \centering
    \includegraphics[width=\linewidth]{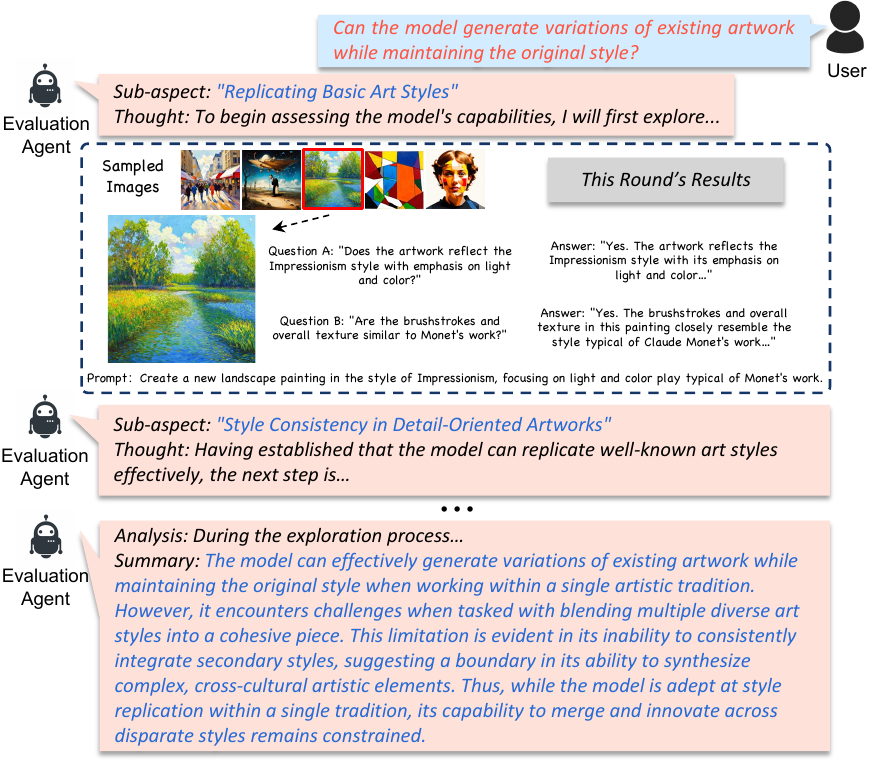}
    \caption{\textbf{A Case of Open-Ended User Query Evaluation.} For open-ended user queries, the Evaluation Agent systematically explores the model’s capabilities in specific areas, starting from basic aspects and gradually delving deeper, culminating in a detailed analysis and summary. Please refer to the Appendix~\ref{open-results} for the complete results.}
    \label{fig:open_case}
    \vspace{-20pt}
\end{figure}

\vspace{-5pt}
\subsection{Experiments on Open-Ended User Query}

We demonstrate the flexibility of our framework and the benefits of its dynamic evaluation through experiments on an open-ended user query dataset that we collect and construct.

\subsubsection{Open-Ended User Query Dataset}

We create an open-ended user query dataset comprising 100 user queries focused on evaluating generative model capabilities. Each query is manually labeled with \texttt{Ability}, \texttt{General/Specific}, and \texttt{Specific Domain} tags. For detailed statistics, please refer to the Appendix~\ref{appendix c}.

\subsubsection{Experimental Setup}
The Evaluation Agent demonstrates strong planning capabilities and accepts any input, but its effectiveness is limited by restrictive evaluation tools, which hinder its ability to handle open-ended queries. To overcome this limitation, we propose a simple yet intuitive solution: leveraging a VLM as an evaluation tool in the form of VQA. During each evaluation round, the PromptGen Agent not only designs prompts for specific sub-aspects but also generates corresponding questions based on the content of each prompt and the aspects to be evaluated. The generated sample and questions are input into the VLM, which provides answers that are then fed back to the Plan Agent for further analysis and planning.

\subsubsection{Open-Ended User Query Evaluation}

Most visual generation benchmarks use predefined dimensions and prompts to evaluate models, but this fixed approach often overlooks users’ specific needs, such as handling unique scenarios or objects. A user study found that 67.44\% (29 of 43) participants prioritized models meeting their specific needs over general performance. Additionally, fixed prompts can lead to targeted optimization, resulting in misleading evaluations.

We address these issues using the Evaluation Agent, which conducts dynamic, multi-round assessments of a model’s capabilities for open-ended queries, with flexible prompt design at each stage. Figure~\ref{fig:open_case} provides an example where a user asks: ``Can the model generate variations of existing artwork while maintaining the original style?'' The agent begins by evaluating the model’s ability to replicate basic artistic styles, confirming its competence. It then narrows its focus to determine whether the model can preserve the original style while modifying specific artwork details. Through iterative evaluations, the agent analyzes and synthesizes the results to provide comprehensive feedback. For the full example, please refer to Appendix~\ref{open-results}. As shown, our Evaluation Agent framework provides precise, detailed, and user-focused evaluations for open-ended queries.

\section{Further Discussions}

In this section, we discuss the unique aspects of the Evaluation Agent compared to traditional benchmarks, as well as its potential broader applications.

\noindent\textbf{Dynamic and Multi-Step Evaluation.} One of the core features of the Evaluation Agent is its dynamic, multi-step evaluation process. This structured evaluation paradigm enables the discovery of nuanced differences in model capabilities and provides a more detailed analysis of a model’s strengths and weaknesses. Specifically, it allows for a hierarchical, step-by-step evaluation, progressing from simple to complex tasks, as well as category-based assessments that measure performance across different content types within the same dimension. In contrast, traditional visual generation evaluation benchmarks, while incorporating diverse prompts carefully designed for each dimension, suffer from limitations such as fixed prompts and a lack of fine-grained prompt categorization. These constraints reduce flexibility, make it harder to draw valuable insights, and increase the risk of models being over-optimized for specific prompts. Furthermore, dynamic evaluations help avoid redundant testing, significantly enhancing efficiency.

\noindent\textbf{Open-Ended Evaluation Toolkit.} Our framework for evaluating open-ended queries relies on two key aspects: the planning and reasoning capabilities of LLM Agents and the evaluation toolkit’s ability to assess diverse dimensions. One approach to building the evaluation toolkit is integrating various tools, allowing the agent to select the most suitable one for each task. However, current evaluation tools are limited, often focusing only on general evaluations and lacking sensitivity to fine-grained details. For instance, CLIPScore~\cite{hessel2021clipscore} captures the general similarity between an image and a caption but fails to detect subtle changes, such as variations in object counts, limiting its effectiveness in specific evaluations. An alternative approach is to design a versatile evaluation tool capable of assessing multiple aspects. A VQA-based format using VLMs is particularly promising, enabling fine-grained evaluation through targeted questions and providing detailed textual outputs that integrate well with LLM Agents. While this method’s effectiveness depends on the VLM’s capabilities, current VLMs already demonstrate impressive results, with future advancements poised to further enhance performance.

\noindent\textbf{Broader Applications.} The Evaluation Agent not only evaluates the performance of a single model but also facilitates the direct comparison of two models’ strengths and weaknesses in specific dimensions by assessing them simultaneously during execution. This feature enables users to determine which model excels in particular areas. Moreover, by accumulating evaluation results across various capabilities, a database can be constructed. Once sufficient information about multiple models is collected, this database can serve as the foundation for building a recommendation system, capable of suggesting the most suitable model based on the user’s specific needs.

\section{Conclusion}
\label{conclusion}

As the first of its kind, our Evaluation Agent redefines how visual generative models can be assessed, moving beyond rigid evaluation pipelines to offer an efficient and promptable approach. Unlike traditional methods that rely on fixed benchmarks and time-consuming sampling processes, the Evaluation Agent mimics human evaluation strategies. This allows for significant reduction in evaluation time while providing flexibility and adaptability to user-specified criteria. Our framework dynamically adjusts the evaluation process, enabling efficient assessments with fewer samples and offering scalable, customizable integration for a wide range of evaluation tools and visual generative models.
By open-sourcing this framework, we aim to inspire further research in the development of more flexible and efficient evaluation methods for visual generative models.

\clearpage
\section{Limitations}

The performance of our Evaluation Agent framework is influenced by two orthogonal factors: 1) the reliability of the Evaluation Toolkit, and 2) the capability of the LLMs used to develop the agentic systems.

\noindent\textbf{Evaluation Toolkit.} While we can integrate state-of-the-art (SOTA) toolkits, they may not always perfectly align with human perception, particularly when evaluating visual generative models. This misalignment can negatively impact the accuracy of evaluation results when the Evaluation Agent relies on these tools. Furthermore, although the Evaluation Agent is designed as an open framework capable of handling arbitrary open-form user queries, existing evaluation tools are still limited in covering certain edge cases or specific criteria that users may want to assess.

\noindent\textbf{LLMs.} We found that even the most advanced LLMs occasionally fall short, such as producing inconsistent output formats or struggling with numerical comparisons. Possible solutions include employing post-processing techniques to refine the outputs and using external tools to handle numerical evaluations. With the release of more powerful models, such as o1, we believe these issues can be largely mitigated.

Our primary contribution is introducing a new evaluation paradigm. We are confident that the utility of the Evaluation Agent framework will continue to improve as stronger LLMs and more human-aligned evaluation toolkits are developed.

\section{Ethical Considerations}

The Evaluation Agent could potentially be used to prompt visual generative models to synthesize unsafe or harmful visual content, such as deepfakes or offensive images and videos. This raises ethical concerns regarding the misuse of these visual generative models. We strongly advise users to approach any system involving visual generative models with caution, as improper use could lead to the creation and spread of harmful content. 

\section{Acknowledgment}
This study is supported by the Ministry of Education, Singapore, under its MOE AcRF Tier 2 (MOET2EP20221-0012, MOE-T2EP20223-0002), and under the RIE2020 Industry Alignment Fund – Industry Collaboration Projects (IAF-ICP) Funding Initiative, as well as cash and in-kind contribution from the industry partner(s).

\clearpage
\bibliography{custom}

\begin{thebibliography}{51}
\providecommand{\natexlab}[1]{#1}

\bibitem[{Bain et~al.(2021)Bain, Nagrani, Varol, and Zisserman}]{Bain21}
Max Bain, Arsha Nagrani, G{\"u}l Varol, and Andrew Zisserman. 2021.
\newblock Frozen in time: A joint video and image encoder for end-to-end retrieval.
\newblock In \emph{IEEE International Conference on Computer Vision}.

\bibitem[{Besta et~al.(2024)Besta, Blach, Kubicek, Gerstenberger, Podstawski, Gianinazzi, Gajda, Lehmann, Niewiadomski, Nyczyk et~al.}]{besta2024graph}
Maciej Besta, Nils Blach, Ales Kubicek, Robert Gerstenberger, Michal Podstawski, Lukas Gianinazzi, Joanna Gajda, Tomasz Lehmann, Hubert Niewiadomski, Piotr Nyczyk, et~al. 2024.
\newblock Graph of thoughts: Solving elaborate problems with large language models.
\newblock In \emph{Proceedings of the AAAI Conference on Artificial Intelligence}, volume~38, pages 17682--17690.

\bibitem[{Chen et~al.(2024{\natexlab{a}})Chen, Zhang, Cun, Xia, Wang, Weng, and Shan}]{chen2024videocrafter2}
Haoxin Chen, Yong Zhang, Xiaodong Cun, Menghan Xia, Xintao Wang, Chao Weng, and Ying Shan. 2024{\natexlab{a}}.
\newblock Videocrafter2: Overcoming data limitations for high-quality video diffusion models.
\newblock \emph{arXiv preprint arXiv:2401.09047}.

\bibitem[{Chen et~al.(2024{\natexlab{b}})Chen, Siarohin, Menapace, Deyneka, Chao, Jeon, Fang, Lee, Ren, Yang, and Tulyakov}]{chen2024panda70m}
Tsai-Shien Chen, Aliaksandr Siarohin, Willi Menapace, Ekaterina Deyneka, Hsiang-wei Chao, Byung~Eun Jeon, Yuwei Fang, Hsin-Ying Lee, Jian Ren, Ming-Hsuan Yang, and Sergey Tulyakov. 2024{\natexlab{b}}.
\newblock Panda-70m: Captioning 70m videos with multiple cross-modality teachers.
\newblock \emph{arXiv preprint arXiv:2402.19479}.

\bibitem[{Chiang and yi~Lee(2023)}]{chiang2023largelanguagemodelsalternative}
Cheng-Han Chiang and Hung yi~Lee. 2023.
\newblock \href {https://arxiv.org/abs/2305.01937} {Can large language models be an alternative to human evaluations?}
\newblock \emph{Preprint}, arXiv:2305.01937.

\bibitem[{Esser et~al.(2024)Esser, Kulal, Blattmann, Entezari, M{\"u}ller, Saini, Levi, Lorenz, Sauer, Boesel et~al.}]{esser2024scaling}
Patrick Esser, Sumith Kulal, Andreas Blattmann, Rahim Entezari, Jonas M{\"u}ller, Harry Saini, Yam Levi, Dominik Lorenz, Axel Sauer, Frederic Boesel, et~al. 2024.
\newblock Scaling rectified flow transformers for high-resolution image synthesis.
\newblock In \emph{Forty-first International Conference on Machine Learning}.

\bibitem[{Fu et~al.(2023)Fu, Ng, Jiang, and Liu}]{fu2023gptscoreevaluatedesire}
Jinlan Fu, See-Kiong Ng, Zhengbao Jiang, and Pengfei Liu. 2023.
\newblock \href {https://arxiv.org/abs/2302.04166} {Gptscore: Evaluate as you desire}.
\newblock \emph{Preprint}, arXiv:2302.04166.

\bibitem[{He et~al.(2022)He, Yang, Zhang, Shan, and Chen}]{he2022lvdm}
Yingqing He, Tianyu Yang, Yong Zhang, Ying Shan, and Qifeng Chen. 2022.
\newblock Latent video diffusion models for high-fidelity video generation with arbitrary lengths.
\newblock \emph{arXiv preprint arXiv:2211.13221}.

\bibitem[{Hessel et~al.(2021)Hessel, Holtzman, Forbes, Bras, and Choi}]{hessel2021clipscore}
Jack Hessel, Ari Holtzman, Maxwell Forbes, Ronan~Le Bras, and Yejin Choi. 2021.
\newblock Clipscore: A reference-free evaluation metric for image captioning.
\newblock \emph{arXiv preprint arXiv:2104.08718}.

\bibitem[{Heusel et~al.(2017)Heusel, Ramsauer, Unterthiner, Nessler, and Hochreiter}]{heusel2017gans}
Martin Heusel, Hubert Ramsauer, Thomas Unterthiner, Bernhard Nessler, and Sepp Hochreiter. 2017.
\newblock Gans trained by a two time-scale update rule converge to a local nash equilibrium.
\newblock \emph{Advances in neural information processing systems}, 30.

\bibitem[{Ho et~al.(2020)Ho, Jain, and Abbeel}]{ho2020denoising}
Jonathan Ho, Ajay Jain, and Pieter Abbeel. 2020.
\newblock Denoising diffusion probabilistic models.
\newblock \emph{arXiv preprint arxiv:2006.11239}.

\bibitem[{Huang et~al.(2023)Huang, Sun, Xie, Li, and Liu}]{huang2023t2icompbench}
Kaiyi Huang, Kaiyue Sun, Enze Xie, Zhenguo Li, and Xihui Liu. 2023.
\newblock T2i-compbench: A comprehensive benchmark for open-world compositional text-to-image generation.
\newblock \emph{arXiv preprint arXiv:2307.06350}.

\bibitem[{Huang et~al.(2024{\natexlab{a}})Huang, He, Yu, Zhang, Si, Jiang, Zhang, Wu, Jin, Chanpaisit, Wang, Chen, Wang, Lin, Qiao, and Liu}]{huang2023vbench}
Ziqi Huang, Yinan He, Jiashuo Yu, Fan Zhang, Chenyang Si, Yuming Jiang, Yuanhan Zhang, Tianxing Wu, Qingyang Jin, Nattapol Chanpaisit, Yaohui Wang, Xinyuan Chen, Limin Wang, Dahua Lin, Yu~Qiao, and Ziwei Liu. 2024{\natexlab{a}}.
\newblock {VBench}: Comprehensive benchmark suite for video generative models.
\newblock In \emph{Proceedings of the IEEE/CVF Conference on Computer Vision and Pattern Recognition}.

\bibitem[{Huang et~al.(2024{\natexlab{b}})Huang, Zhang, Xu, He, Yu, Dong, Ma, Chanpaisit, Si, Jiang, Wang, Chen, Chen, Wang, Lin, Qiao, and Liu}]{huang2024vbench++}
Ziqi Huang, Fan Zhang, Xiaojie Xu, Yinan He, Jiashuo Yu, Ziyue Dong, Qianli Ma, Nattapol Chanpaisit, Chenyang Si, Yuming Jiang, Yaohui Wang, Xinyuan Chen, Ying-Cong Chen, Limin Wang, Dahua Lin, Yu~Qiao, and Ziwei Liu. 2024{\natexlab{b}}.
\newblock Vbench++: Comprehensive and versatile benchmark suite for video generative models.
\newblock \emph{arXiv preprint arXiv:2411.13503}.

\bibitem[{Jain et~al.(2023)Jain, Keshava, Mysore~Sathyendra, Fernandes, Liu, Neubig, and Zhou}]{Jain_2023}
Sameer Jain, Vaishakh Keshava, Swarnashree Mysore~Sathyendra, Patrick Fernandes, Pengfei Liu, Graham Neubig, and Chunting Zhou. 2023.
\newblock \href {https://doi.org/10.18653/v1/2023.findings-acl.537} {Multi-dimensional evaluation of text summarization with in-context learning}.
\newblock In \emph{Findings of the Association for Computational Linguistics: ACL 2023}, page 8487–8495. Association for Computational Linguistics.

\bibitem[{Kapoor et~al.(2024)Kapoor, Butala, Russak, Koh, Kamble, Alshikh, and Salakhutdinov}]{kapoor2024omniact}
Raghav Kapoor, Yash~Parag Butala, Melisa Russak, Jing~Yu Koh, Kiran Kamble, Waseem Alshikh, and Ruslan Salakhutdinov. 2024.
\newblock Omniact: A dataset and benchmark for enabling multimodal generalist autonomous agents for desktop and web.
\newblock \emph{arXiv preprint arXiv:2402.17553}.

\bibitem[{Kojima et~al.(2022)Kojima, Gu, Reid, Matsuo, and Iwasawa}]{kojima2022large}
Takeshi Kojima, Shixiang~Shane Gu, Machel Reid, Yutaka Matsuo, and Yusuke Iwasawa. 2022.
\newblock Large language models are zero-shot reasoners.
\newblock \emph{Advances in neural information processing systems}, 35:22199--22213.

\bibitem[{Lee et~al.(2023)Lee, Yasunaga, Meng, Mai, Park, Gupta, Zhang, Narayanan, Teufel, Bellagente et~al.}]{lee2023holistic}
Tony Lee, Michihiro Yasunaga, Chenlin Meng, Yifan Mai, Joon~Sung Park, Agrim Gupta, Yunzhi Zhang, Deepak Narayanan, Hannah Teufel, Marco Bellagente, et~al. 2023.
\newblock Holistic evaluation of text-to-image models.
\newblock \emph{Advances in Neural Information Processing Systems}, 36:69981--70011.

\bibitem[{Li et~al.(2023{\natexlab{a}})Li, Zhao, Yu, Song, Li, Yu, Li, Huang, and Li}]{li2023api}
Minghao Li, Yingxiu Zhao, Bowen Yu, Feifan Song, Hangyu Li, Haiyang Yu, Zhoujun Li, Fei Huang, and Yongbin Li. 2023{\natexlab{a}}.
\newblock Api-bank: A comprehensive benchmark for tool-augmented llms.
\newblock \emph{arXiv preprint arXiv:2304.08244}.

\bibitem[{Li et~al.(2023{\natexlab{b}})Li, Cui, Kong, and Bi}]{li2023collaborativeevaluationexploringsynergy}
Qintong Li, Leyang Cui, Lingpeng Kong, and Wei Bi. 2023{\natexlab{b}}.
\newblock \href {https://arxiv.org/abs/2310.19740} {Collaborative evaluation: Exploring the synergy of large language models and humans for open-ended generation evaluation}.
\newblock \emph{Preprint}, arXiv:2310.19740.

\bibitem[{Liu et~al.(2023)Liu, Cun, Liu, Wang, Zhang, Chen, Liu, Zeng, Chan, and Shan}]{liu2023evalcrafter}
Yaofang Liu, Xiaodong Cun, Xuebo Liu, Xintao Wang, Yong Zhang, Haoxin Chen, Yang Liu, Tieyong Zeng, Raymond Chan, and Ying Shan. 2023.
\newblock Evalcrafter: Benchmarking and evaluating large video generation models.
\newblock \emph{arXiv preprint arXiv:2310.11440}.

\bibitem[{Ma et~al.(2024)Ma, Wang, Jia, Chen, Liu, Li, Chen, and Qiao}]{ma2024latte}
Xin Ma, Yaohui Wang, Gengyun Jia, Xinyuan Chen, Ziwei Liu, Yuan-Fang Li, Cunjian Chen, and Yu~Qiao. 2024.
\newblock Latte: Latent diffusion transformer for video generation.
\newblock \emph{arXiv preprint arXiv:2401.03048}.

\bibitem[{Miao et~al.(2023)Miao, Teh, and Rainforth}]{miao2023selfcheck}
Ning Miao, Yee~Whye Teh, and Tom Rainforth. 2023.
\newblock Selfcheck: Using llms to zero-shot check their own step-by-step reasoning.
\newblock \emph{arXiv preprint arXiv:2308.00436}.

\bibitem[{Pan et~al.(2024)Pan, Zhang, Tomlin, Zhou, Levine, and Suhr}]{pan2024autonomousevaluationrefinementdigital}
Jiayi Pan, Yichi Zhang, Nicholas Tomlin, Yifei Zhou, Sergey Levine, and Alane Suhr. 2024.
\newblock \href {https://arxiv.org/abs/2404.06474} {Autonomous evaluation and refinement of digital agents}.
\newblock \emph{Preprint}, arXiv:2404.06474.

\bibitem[{Patil et~al.(2023)Patil, Zhang, Wang, and Gonzalez}]{patil2023gorillalargelanguagemodel}
Shishir~G. Patil, Tianjun Zhang, Xin Wang, and Joseph~E. Gonzalez. 2023.
\newblock \href {https://arxiv.org/abs/2305.15334} {Gorilla: Large language model connected with massive apis}.
\newblock \emph{Preprint}, arXiv:2305.15334.

\bibitem[{Podell et~al.(2023)Podell, English, Lacey, Blattmann, Dockhorn, M{\"u}ller, Penna, and Rombach}]{podell2023sdxl}
Dustin Podell, Zion English, Kyle Lacey, Andreas Blattmann, Tim Dockhorn, Jonas M{\"u}ller, Joe Penna, and Robin Rombach. 2023.
\newblock Sdxl: Improving latent diffusion models for high-resolution image synthesis.
\newblock \emph{arXiv preprint arXiv:2307.01952}.

\bibitem[{Qin et~al.(2023)Qin, Liang, Ye, Zhu, Yan, Lu, Lin, Cong, Tang, Qian et~al.}]{qin2023toolllm}
Yujia Qin, Shihao Liang, Yining Ye, Kunlun Zhu, Lan Yan, Yaxi Lu, Yankai Lin, Xin Cong, Xiangru Tang, Bill Qian, et~al. 2023.
\newblock Toolllm: Facilitating large language models to master 16000+ real-world apis.
\newblock \emph{arXiv preprint arXiv:2307.16789}.

\bibitem[{Qin et~al.(2024)Qin, Cheng, Wang, Yi, Shao, Fan, Li, and Lao}]{qin2024evaluating}
Ziyuan Qin, Dongjie Cheng, Haoyu Wang, Huahui Yi, Yuting Shao, Zhiyuan Fan, Kang Li, and Qicheng Lao. 2024.
\newblock Evaluating hallucination in text-to-image diffusion models with scene-graph based question-answering agent.
\newblock \emph{arXiv preprint arXiv:2412.05722}.

\bibitem[{Rombach et~al.(2022)Rombach, Blattmann, Lorenz, Esser, and Ommer}]{rombach2022high}
Robin Rombach, Andreas Blattmann, Dominik Lorenz, Patrick Esser, and Bj{\"o}rn Ommer. 2022.
\newblock High-resolution image synthesis with latent diffusion models.
\newblock In \emph{Proceedings of the IEEE/CVF conference on computer vision and pattern recognition}, pages 10684--10695.

\bibitem[{Schick et~al.(2024)Schick, Dwivedi-Yu, Dess{\`\i}, Raileanu, Lomeli, Hambro, Zettlemoyer, Cancedda, and Scialom}]{schick2024toolformer}
Timo Schick, Jane Dwivedi-Yu, Roberto Dess{\`\i}, Roberta Raileanu, Maria Lomeli, Eric Hambro, Luke Zettlemoyer, Nicola Cancedda, and Thomas Scialom. 2024.
\newblock Toolformer: Language models can teach themselves to use tools.
\newblock \emph{Advances in Neural Information Processing Systems}, 36.

\bibitem[{Sel et~al.(2023)Sel, Al-Tawaha, Khattar, Jia, and Jin}]{sel2023algorithm}
Bilgehan Sel, Ahmad Al-Tawaha, Vanshaj Khattar, Ruoxi Jia, and Ming Jin. 2023.
\newblock Algorithm of thoughts: Enhancing exploration of ideas in large language models.
\newblock \emph{arXiv preprint arXiv:2308.10379}.

\bibitem[{Shen et~al.(2024)Shen, Song, Tan, Li, Lu, and Zhuang}]{shen2024hugginggpt}
Yongliang Shen, Kaitao Song, Xu~Tan, Dongsheng Li, Weiming Lu, and Yueting Zhuang. 2024.
\newblock Hugginggpt: Solving ai tasks with chatgpt and its friends in hugging face.
\newblock \emph{Advances in Neural Information Processing Systems}, 36.

\bibitem[{Shinn et~al.(2024)Shinn, Cassano, Gopinath, Narasimhan, and Yao}]{shinn2024reflexion}
Noah Shinn, Federico Cassano, Ashwin Gopinath, Karthik Narasimhan, and Shunyu Yao. 2024.
\newblock Reflexion: Language agents with verbal reinforcement learning.
\newblock \emph{Advances in Neural Information Processing Systems}, 36.

\bibitem[{Sun et~al.(2024)Sun, Huang, Liu, Wu, Xu, Li, and Liu}]{sun2024t2v}
Kaiyue Sun, Kaiyi Huang, Xian Liu, Yue Wu, Zihan Xu, Zhenguo Li, and Xihui Liu. 2024.
\newblock T2v-compbench: A comprehensive benchmark for compositional text-to-video generation.
\newblock \emph{arXiv preprint arXiv:2407.14505}.

\bibitem[{Unterthiner et~al.(2018)Unterthiner, Van~Steenkiste, Kurach, Marinier, Michalski, and Gelly}]{unterthiner2018towards}
Thomas Unterthiner, Sjoerd Van~Steenkiste, Karol Kurach, Raphael Marinier, Marcin Michalski, and Sylvain Gelly. 2018.
\newblock Towards accurate generative models of video: A new metric \& challenges.
\newblock \emph{arXiv preprint arXiv:1812.01717}.

\bibitem[{Wang et~al.(2023)Wang, Yuan, Chen, Zhang, Wang, and Zhang}]{wang2023modelscope}
Jiuniu Wang, Hangjie Yuan, Dayou Chen, Yingya Zhang, Xiang Wang, and Shiwei Zhang. 2023.
\newblock Modelscope text-to-video technical report.
\newblock \emph{arXiv preprint arXiv:2308.06571}.

\bibitem[{Wang et~al.(2024{\natexlab{a}})Wang, Xu, Ye, Yan, Shen, Zhang, Huang, and Sang}]{wang2024mobile}
Junyang Wang, Haiyang Xu, Jiabo Ye, Ming Yan, Weizhou Shen, Ji~Zhang, Fei Huang, and Jitao Sang. 2024{\natexlab{a}}.
\newblock Mobile-agent: Autonomous multi-modal mobile device agent with visual perception.
\newblock \emph{arXiv preprint arXiv:2401.16158}.

\bibitem[{Wang et~al.(2024{\natexlab{b}})Wang, Ma, Feng, Zhang, Yang, Zhang, Chen, Tang, Chen, Lin et~al.}]{wang2024survey}
Lei Wang, Chen Ma, Xueyang Feng, Zeyu Zhang, Hao Yang, Jingsen Zhang, Zhiyuan Chen, Jiakai Tang, Xu~Chen, Yankai Lin, et~al. 2024{\natexlab{b}}.
\newblock A survey on large language model based autonomous agents.
\newblock \emph{Frontiers of Computer Science}, 18(6):186345.

\bibitem[{Wang et~al.(2022)Wang, Wei, Schuurmans, Le, Chi, Narang, Chowdhery, and Zhou}]{wang2022self}
Xuezhi Wang, Jason Wei, Dale Schuurmans, Quoc Le, Ed~Chi, Sharan Narang, Aakanksha Chowdhery, and Denny Zhou. 2022.
\newblock Self-consistency improves chain of thought reasoning in language models.
\newblock \emph{arXiv preprint arXiv:2203.11171}.

\bibitem[{Wei et~al.(2022)Wei, Wang, Schuurmans, Bosma, Xia, Chi, Le, Zhou et~al.}]{wei2022chain}
Jason Wei, Xuezhi Wang, Dale Schuurmans, Maarten Bosma, Fei Xia, Ed~Chi, Quoc~V Le, Denny Zhou, et~al. 2022.
\newblock Chain-of-thought prompting elicits reasoning in large language models.
\newblock \emph{Advances in neural information processing systems}, 35:24824--24837.

\bibitem[{Xi et~al.(2023)Xi, Chen, Guo, He, Ding, Hong, Zhang, Wang, Jin, Zhou et~al.}]{xi2023rise}
Zhiheng Xi, Wenxiang Chen, Xin Guo, Wei He, Yiwen Ding, Boyang Hong, Ming Zhang, Junzhe Wang, Senjie Jin, Enyu Zhou, et~al. 2023.
\newblock The rise and potential of large language model based agents: A survey.
\newblock \emph{arXiv preprint arXiv:2309.07864}.

\bibitem[{Xie et~al.(2024)Xie, Zhang, Chen, Li, Zhao, Cao, Hua, Cheng, Shin, Lei et~al.}]{xie2024osworld}
Tianbao Xie, Danyang Zhang, Jixuan Chen, Xiaochuan Li, Siheng Zhao, Ruisheng Cao, Toh~Jing Hua, Zhoujun Cheng, Dongchan Shin, Fangyu Lei, et~al. 2024.
\newblock Osworld: Benchmarking multimodal agents for open-ended tasks in real computer environments.
\newblock \emph{arXiv preprint arXiv:2404.07972}.

\bibitem[{Xue et~al.(2019)Xue, Chen, Wu, Wei, and Freeman}]{xue2019video}
Tianfan Xue, Baian Chen, Jiajun Wu, Donglai Wei, and William~T Freeman. 2019.
\newblock Video enhancement with task-oriented flow.
\newblock \emph{International Journal of Computer Vision (IJCV)}, 127(8):1106--1125.

\bibitem[{Yang et~al.(2025)Yang, Fan, Sun, Li, Zeng, Han, Zhai, Liu, Cao, and Zha}]{yang2025videogen}
Yuhang Yang, Ke~Fan, Shangkun Sun, Hongxiang Li, Ailing Zeng, FeiLin Han, Wei Zhai, Wei Liu, Yang Cao, and Zheng-Jun Zha. 2025.
\newblock Videogen-eval: Agent-based system for video generation evaluation.
\newblock \emph{arXiv preprint arXiv:2503.23452}.

\bibitem[{Yao et~al.(2024)Yao, Yu, Zhao, Shafran, Griffiths, Cao, and Narasimhan}]{yao2024tree}
Shunyu Yao, Dian Yu, Jeffrey Zhao, Izhak Shafran, Tom Griffiths, Yuan Cao, and Karthik Narasimhan. 2024.
\newblock Tree of thoughts: Deliberate problem solving with large language models.
\newblock \emph{Advances in Neural Information Processing Systems}, 36.

\bibitem[{Yao et~al.(2022)Yao, Zhao, Yu, Du, Shafran, Narasimhan, and Cao}]{yao2022react}
Shunyu Yao, Jeffrey Zhao, Dian Yu, Nan Du, Izhak Shafran, Karthik Narasimhan, and Yuan Cao. 2022.
\newblock React: Synergizing reasoning and acting in language models.
\newblock \emph{arXiv preprint arXiv:2210.03629}.

\bibitem[{Zhang et~al.(2024{\natexlab{a}})Zhang, Yuan, and Yao}]{zhang2024diagram}
Yifan Zhang, Yang Yuan, and Andrew Chi-Chih Yao. 2024{\natexlab{a}}.
\newblock On the diagram of thought.
\newblock \emph{arXiv preprint arXiv:2409.10038}.

\bibitem[{Zhang et~al.(2024{\natexlab{b}})Zhang, Tian, Chen, and Liu}]{zhang2024mmina}
Ziniu Zhang, Shulin Tian, Liangyu Chen, and Ziwei Liu. 2024{\natexlab{b}}.
\newblock Mmina: Benchmarking multihop multimodal internet agents.
\newblock \emph{arXiv preprint arXiv:2404.09992}.

\bibitem[{Zheng et~al.(2025)Zheng, Huang, Liu, Zou, He, Zhang, Zhang, He, Zheng, Qiao, and Liu}]{zheng2025vbench2}
Dian Zheng, Ziqi Huang, Hongbo Liu, Kai Zou, Yinan He, Fan Zhang, Yuanhan Zhang, Jingwen He, Wei-Shi Zheng, Yu~Qiao, and Ziwei Liu. 2025.
\newblock {VBench-2.0}: Advancing video generation benchmark suite for intrinsic faithfulness.
\newblock \emph{arXiv preprint arXiv:2503.21755}.

\bibitem[{Zheng et~al.(2023)Zheng, Chiang, Sheng, Zhuang, Wu, Zhuang, Lin, Li, Li, Xing et~al.}]{zheng2023judging}
Lianmin Zheng, Wei-Lin Chiang, Ying Sheng, Siyuan Zhuang, Zhanghao Wu, Yonghao Zhuang, Zi~Lin, Zhuohan Li, Dacheng Li, Eric Xing, et~al. 2023.
\newblock Judging llm-as-a-judge with mt-bench and chatbot arena.
\newblock \emph{Advances in Neural Information Processing Systems}, 36:46595--46623.

\bibitem[{Zhou et~al.(2023)Zhou, Xu, Zhu, Zhou, Lo, Sridhar, Cheng, Ou, Bisk, Fried et~al.}]{zhou2023webarena}
Shuyan Zhou, Frank~F Xu, Hao Zhu, Xuhui Zhou, Robert Lo, Abishek Sridhar, Xianyi Cheng, Tianyue Ou, Yonatan Bisk, Daniel Fried, et~al. 2023.
\newblock Webarena: A realistic web environment for building autonomous agents.
\newblock \emph{arXiv preprint arXiv:2307.13854}.

\end{thebibliography}

\clearpage
\appendix
\section*{Supplementary}

In this \textbf{\textit{supplementary file}}, we provide a detailed explanation of the pipeline in Section~\ref{appendix a}. Next, we present additional experimental details in Section~\ref{appendix b} and elaborate on the open-ended user query dataset in Section~\ref{appendix c}. Furthermore, we discuss additional related work in Section~\ref{appendix d}. Finally, in Section~\ref{appendix e}, we present further experimental results and analyses using different base models, along with a variety of comprehensive evaluation results for open-ended user queries.

\section{Detailed Explanation of Pipeline}
\label{appendix a}

Our dynamic evaluation pipeline consists of two stages: the Proposal Stage and the Execution Stage. By iteratively interacting and looping between these stages, the framework dynamically evaluates models in response to user queries.

\subsection{Proposal Stage}

The Proposal Stage consists of two agents: the Plan Agent and the PromptGen Agent. The Plan Agent is responsible for planning each step and providing the final summary and analysis, while the PromptGen Agent specializes in designing prompts for the process.

\noindent\textbf{Plan Agent.}
When humans are interested in a specific aspect of a model’s capabilities, they often generate content to observe its performance. Through several rounds of iterative generation and observation, they can form a preliminary evaluation of the model’s ability in that aspect. During this process, before each round of generation, humans typically consider which direction to focus on. We designed the Plan Agent to simulate this decision-making behavior.

The Plan Agent primarily simulates the decision-making process in human evaluations. It is responsible for determining the direction of exploration at each step and for summarizing and analyzing the results. Specifically, at the beginning, the Plan Agent receives a user query related to the model’s capabilities. We require the Plan Agent to first propose an initial aspect to explore based on this query. In each subsequent step, it needs to consider both the user’s original query and the observations from intermediate results to suggest further directions for exploration, until sufficient information is gathered to evaluate the model’s capability in relation to the query. When the Plan Agent believes it has gathered sufficient information, it will analyze and summarize all the observed results from the exploration process, and then return the final evaluation results to the human.

In this process, we require the Plan Agent to provide an explanation (thought) for why it chooses a particular direction to explore at each step. When it chooses to respond to the human, it is also required to explain why it believes sufficient information has been gathered to stop further exploration.

For tool usage, in the case of closed-domain queries that involve experiments on existing benchmarks, the Plan Agent is required to not only propose the direction to explore at each step but also specify the appropriate evaluation tool for that direction. For open-ended queries, we employ a Vision-Language Model (VLM) for evaluation, using the Visual Question Answering (VQA) format.

\noindent\textbf{PromptGen Agent.}
Once the human determines the direction to focus on, corresponding prompts must be designed based on that focus for model sampling and generation. To simulate this behavior, we developed the PromptGen Agent. Specifically, at each step, the PromptGen Agent receives the “aspect to explore” proposed by the Plan Agent and designs prompts aligned with the specified exploration direction. Additionally, while crafting each prompt, the PromptGen Agent provides an explanation of the reasoning behind its design.

\subsection{Execution Stage}
The Execution Stage is responsible for sampling and evaluating the visual generation model based on the evaluation tools selected by the Plan Agent and the prompts designed by the PromptGen Agent.

\noindent\textbf{Sampling and Evaluation.}
For the designed prompts, we use a visual generation model to sample and generate corresponding images or videos, and then we need to evaluate the quality of the generated content. For humans, the evaluation process is essentially an observation of the generated content. In our pipeline, for closed-domain questions, we evaluate specific dimensions by invoking existing evaluation tools within the closed domain. For open-ended queries, we use a VLM to simulate human evaluation of the generated content in the form of VQA. Finally, we integrate the evaluation results of each generated piece and return them to the Plan Agent for observation and analysis.

\subsection{Dynamic Looping.}
Dynamic evaluation refers to initially providing a preliminary focus based on the user’s query, and then continuously adjusting what aspects to focus on according to the intermediate evaluation results. This can involve adjustments in terms of a wider variety of scenarios, more complex situations, or more intricate prompts, among other factors, until the Plan Agent believe it have gathered enough information to answer the user’s original query.

\begin{figure*}[htbp!]
    \centering
    \includegraphics[width=0.9\textwidth]{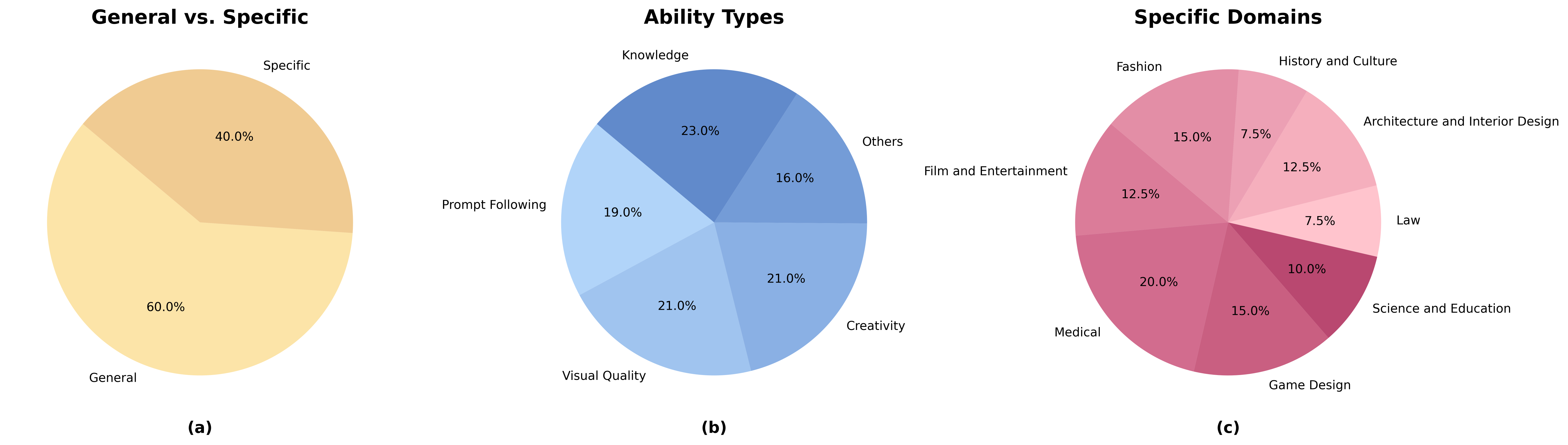}
    \caption{\textbf{Data Distribution of Open-Ended User Query Dataset.} We analyze the constructed open-ended user query dataset from three aspects: General/Specific, Ability, and Specific Domain. The results indicate that our dataset exhibits a relatively balanced distribution across these dimensions.}
    \label{fig:open_domain_stat}
    \vspace{-5pt}
\end{figure*}

\begin{table*}[htbp]
\captionof{table}{\textbf{Evaluation Results Comparison with VBench~\cite{huang2023vbench} using Claude as Base Model}. We adhere to the same experimental settings and parameters as in the main experiments, but we replace the planning and reasoning agents’ backbones with \texttt{claude-3-5-sonnet-20241022} as the base model.}
    \vspace{-5pt}
    \centering
    \small
    \setlength{\tabcolsep}{2pt}
    \resizebox{\textwidth}{!}{%
        \begin{tabular}{>{\centering\arraybackslash}p{4.5cm}|
                            >{\centering\arraybackslash}p{2cm}|
                            >{\centering\arraybackslash}p{2cm}|
                            >{\centering\arraybackslash}p{2cm}|
                            >{\centering\arraybackslash}p{2cm}|
                            >{\centering\arraybackslash}p{2cm}|
                            >{\centering\arraybackslash}p{2cm}|
                            >{\centering\arraybackslash}p{2cm}}
        \Xhline{1pt}
        \textbf{Models} & \textbf{\Centerstack{Subject\\Consistency}} & \textbf{\Centerstack{Background\\Consistency}} & 
        \textbf{\Centerstack{Motion\\Smoothness}} & 
        \textbf{\Centerstack{Dynamic\\Degree}} & 
        \textbf{\Centerstack{Aesthetic\\Quality}} & 
        \textbf{\Centerstack{Imaging\\Quality}} & 
        \textbf{\Centerstack{Object\\Class}} \\
        \Xhline{1pt}
        Latte-1~\cite{ma2024latte} & 0{\%} / 10{\%} & 0{\%} / 10{\%} & 0{\%} / 30{\%} & 0{\%} / 40{\%} & 100{\%} / 100{\%} & 90{\%} / 100{\%} & 0{\%} / 30{\%} \\ 
        ModelScope~\cite{wang2023modelscope}   & 0{\%} / 10{\%} & 30{\%} / 40{\%} & 10{\%} / 80{\%} & 30{\%} / 100{\%} & 40{\%} / 100{\%} & 60{\%} / 100{\%} & 20{\%} / 50{\%} \\ 
        VideoCrafter-0.9~\cite{he2022lvdm}  & 40{\%} / 100{\%} & 30{\%} / 80{\%} & 40{\%} / 90{\%} & 90{\%} / 100{\%} & 90{\%} / 100{\%} & 20{\%} / 100{\%} & 10{\%} / 40{\%} \\ 
        VideoCrafter-2~\cite{chen2024videocrafter2}  & 50{\%} / 100{\%} & 0{\%} / 100{\%} & 0{\%} / 10{\%} & 60{\%} / 100{\%} & 100{\%} / 100{\%} & 80{\%} / 100{\%} & 60{\%} / 90{\%} \\ 
        \Xhline{1pt}    
        \end{tabular}
    }

    \vspace{5pt}

    \setlength{\tabcolsep}{2pt}
    \resizebox{\textwidth}{!}{%
    \begin{tabular}{>{\centering\arraybackslash}p{4.5cm}|
                            >{\centering\arraybackslash}p{2cm}|
                            >{\centering\arraybackslash}p{2cm}|
                            >{\centering\arraybackslash}p{2cm}|
                            >{\centering\arraybackslash}p{2cm}|
                            >{\centering\arraybackslash}p{2cm}|
                            >{\centering\arraybackslash}p{2cm}|
                            >{\centering\arraybackslash}p{2cm}}
    \Xhline{1pt}
    \textbf{\Centerstack{Multiple\\Objects}} &
    \textbf{\Centerstack{Human\\Action}} & 
    \textbf{\Centerstack{Color}} & 
    \textbf{\Centerstack{Spatial\\Relationship}} & 
    \textbf{\Centerstack{Scene}} & 
    \textbf{\Centerstack{Appearance\\Style}} & 
    \textbf{\Centerstack{Temporal\\Style}} & 
    \textbf{\Centerstack{Overall\\Consistency}} \\
    \Xhline{1pt}
    10{\%} / 60{\%} & 60{\%} / 70{\%} & 10{\%} / 60{\%} & 30{\%} / 80{\%} & 0{\%} / 40{\%} & 30{\%} / 100{\%} & 80{\%} / 100{\%} & 80{\%} / 100{\%} \\
    90{\%} / 100{\%} & 40{\%} / 90{\%} & 10{\%} / 20{\%} & 50{\%} / 80{\%} & 40{\%} / 100{\%} & 70{\%} / 100{\%} & 90{\%} / 100{\%} & 40{\%} / 100{\%} \\

    0{\%} / 40{\%} & 20{\%} / 40{\%} & 10{\%} / 40{\%} & 40{\%} / 100{\%} & 10{\%} / 80{\%} & 100{\%} / 100{\%} & 90{\%} / 100{\%} & 60{\%} / 100{\%} \\
    50{\%} / 100{\%} & 50{\%} / 80{\%} & 60{\%} / 90{\%} & 50{\%} / 100{\%} & 0{\%} / 50{\%} & 10{\%} / 100{\%} & 80{\%} / 100{\%} & 80{\%} / 100{\%} \\
    \Xhline{1pt}
    \end{tabular}
    }
    \label{tab:abl_results_t2v_claude}
\end{table*}

\begin{table*}[htbp]
    \caption{\textbf{Evaluation Results Comparison with T2I-CompBench~\cite{huang2023t2icompbench} using Claude as Base Model}. We follow the same experimental setting and the parameters in the main experiments but changing the planning and reasoning agent's backbones with \texttt{claude-3-5-sonnet-20241022} as the base model.} 
    \vspace{-5pt}
    \centering
    \small
    \setlength{\tabcolsep}{3pt}
    \begin{tabular}{c|c|c|c|c}
    \Xhline{1pt}
    \textbf{Models} & \textbf{\Centerstack{Color\\Binding}} &     \textbf{\Centerstack{Shape\\Binding}} &
    \textbf{\Centerstack{Texture\\Binding}} &     \textbf{\Centerstack{Non-Spatial\\Relationships}} \\ \Xhline{1pt}
    SD1.4~\cite{rombach2022high}        & 80{\%} / 100{\%} & 70{\%} / 100{\%} & 80{\%} / 100{\%} & 70{\%} / 100{\%} \\ 
    SD2.1 ~\cite{rombach2022high}  & 80{\%} / 100{\%} & 30{\%} / 100{\%} & 60{\%} / 100{\%} & 70{\%} / 100{\%} \\ 
    SDXL ~\cite{podell2023sdxl}  & 90{\%} / 100{\%} & 60{\%} / 100{\%} & 70{\%} / 100{\%} & 30{\%} / 100{\%} \\ 
    SD3.0 ~\cite{esser2024scaling}  & 10{\%} / 100{\%} & 20{\%} / 100{\%} & 30{\%} / 100{\%} & 20{\%} / 100{\%} \\ 
    \Xhline{1pt}
    \end{tabular}
    \label{tab:abl_results_t2i_claude}
\end{table*}

\begin{table*}[htbp]
    \centering
    \setlength\tabcolsep{3pt}
    \small
    \caption{\textbf{Time Cost Comparison across Models for VBench~\cite{huang2023vbench} Dimensions using Claude as Base Model.} This table compares the evaluation time of four different models using the original VBench pipelines versus the Evaluation Agent. The Evaluation Agent significantly reduces the overall evaluation time.}
    \vspace{-5pt}
    \resizebox{0.99\linewidth}{!}{
    \begin{tabular}{c||c|c||c}
        \Xhline{1pt}
        \textbf{Models}   & \textbf{\Centerstack{VBench (Total Cost) $ \downarrow $}} & \textbf{\Centerstack{VBench (Avg. Cost per Dimension) $ \downarrow $}} & \textbf{\Centerstack{Evaluation Agent (Ours) $ \downarrow $}} \\ \Xhline{1pt}
        Latte-1~\cite{ma2024latte} & 2557 min, 4355 samples & 170 min, 290 samples & 12 min, 15 samples  \\ 
        ModelScope~\cite{wang2023modelscope} & 1160 min, 4355 samples & 77 min, 290 samples & 9 min, 16 samples  \\ 
        VideoCrafter-0.9~\cite{he2022lvdm}  & 1459 min, 4355 samples & 97 min, 290 samples & 9 min, 12 samples  \\ 
        VideoCrafter-2~\cite{chen2024videocrafter2}  & 4261 min, 4355 samples & 284 min, 290 samples & 26 min, 11 samples  \\  \hline
    \end{tabular}
    }
    \label{tab:t2v_time_claude}
\end{table*}

\begin{table*}[htbp]
    \centering
    \setlength\tabcolsep{3pt}
    \small
    \caption{\textbf{Time Cost Comparison across Models for T2I-CompBench~\cite{huang2023t2icompbench} Dimensions using Claude as Base Model.} This table compares the evaluation costs for assessing four models across T2I-CompBench dimensions using both the original T2I-CompBench pipelines and our Evaluation Agent. The Evaluation Agent achieves a substantial reduction in evaluation time compared to the traditional pipelines.
    }
    \vspace{-5pt}
    \resizebox{0.99\linewidth}{!}{
    \begin{tabular}{c||c|c||c}
        \Xhline{1pt}
        \textbf{Models}   & \textbf{\Centerstack{T2I-Comp (Total Cost) $ \downarrow $}} & \textbf{\Centerstack{T2I-Comp (Avg. Cost per Dimension) $ \downarrow $}} & \textbf{\Centerstack{Evaluation Agent (Ours) $ \downarrow $}} \\ \Xhline{1pt}
        SD1.4~\cite{rombach2022high} & 563 min, 12000 samples & 141 min, 3000 samples & 6 min, 18 samples  \\ 
        SD2.1~\cite{rombach2022high} & 782 min, 12000 samples & 196 min, 3000 samples & 7 min, 17 samples  \\ 
        SDXL~\cite{podell2023sdxl}  & 1543 min, 12000 samples & 386 min, 3000 samples & 12 min, 14 samples  \\ 
        SD3.0~\cite{esser2024scaling}  & 1410 min, 12000 samples & 353 min, 3000 samples & 16 min, 13 samples  \\  \hline
    \end{tabular}
    }
    \label{tab:t2i_time_claude}
\end{table*}

\section{Experiment Implementation Details}
\label{appendix b}

All experiments in the main text were implemented using LLMs as the backbone, with \texttt{gpt-4o-2024-08-06} as the core model set to a temperature of 0.7. The system prompt design was inspired by the CoT~\cite{wei2022chain} and ReAct~\cite{yao2022react} frameworks, guiding the agent to solve problems step-by-step and provide explanations at each stage.

For T2V tasks, we validate our evaluation approach by comparing the consistency of our Evaluation Agent with VBench’s original evaluation scheme across multiple dimensions, with respect to evaluation time, sample count, and final assessment results on four open-source T2V models. For VBench, we selected 15 evaluation dimensions, including \texttt{Subject Consistency}, \texttt{Background Consistency}, and \texttt{Motion Smoothness}, among others. Details could be referred to in Table~\ref{tab:main_results_t2v}. We used VBench’s original scheme to sample and evaluate four models on these dimensions, recording generation time, evaluation time, and results. To support a structured evaluation, we implemented a unified scoring pipeline aligned with VBench’s benchmark standards. Specifically, for each evaluation dimension, we first used the official metric tools to score a large number of video samples and partitioned the resulting sample-level scores into five predefined tiers: \texttt{Very High}, \texttt{High}, \texttt{Moderate}, \texttt{Low}, and \texttt{Very Low}, based on the score density distributions. For dimensions scored by proportions, such as \texttt{Dynamic Degree}, we constructed tiering thresholds based on results from $42$ models on VBench’s leaderboard. To ensure fairness, the Evaluation Agent was restricted to the tools and prompts provided by VBench. During evaluation, the agent answered targeted questions, invoked appropriate tools, and selected prompts accordingly. For each sampled video, it executed the relevant metric function to obtain a numerical score, which was then mapped into one of the five tiers using the predefined criteria. Over multiple rounds of evaluation, the agent aggregated these tiered results and determined the model’s overall performance level for each dimension. This process effectively translates intermediate quantitative observations into a final categorical assessment.


For T2I tasks, we conducted the experiment following the similar settings used for T2V models. For T2I-CompBench, we selected $4$ dimensions for the experiment, which are: \texttt{{Color Binding, Shape Binding, Texture Binding, Non-Spatial Relationships}}. Using T2I-CompBench’s original evaluation scheme and prompt list, we evaluated the $4$ models and categorized scores into $5$ tiers, as in the T2V experiment. We excluded the \texttt{Spatial Relationships} dimension due to its reliance on statistical values and limited sample distribution, along with the absence of leaderboard-based references for tiering. The Evaluation Agent was restricted to using the evaluation tools and prompts for these four dimensions and was provided with detailed definitions and tiered results for accurate assessment.

\section{Open-Ended User Query Dataset}
\label{appendix c}

\subsection{Building an Open-Ended User Query Dataset.} To create a dataset of user queries focused on evaluating generative model capabilities, we conducted a user study, gathering user queries from various sources about the aspects users find most important when assessing new models. After cleaning, filtering, and expanding the initial collection, we compiled a dataset of 100 open-ended user queries. 

\subsection{Dataset Statistics.}
We manually annotated each query with labels for \texttt{Ability}, \texttt{General/Specific}, and \texttt{Specific Domain} for analysis. The \texttt{Ability} label categorizes the model capabilities targeted by the question into five types: \texttt{Prompt Following}, \texttt{Visual Quality}, \texttt{Creativity}, \texttt{Knowledge}, and \texttt{Others}. For the \texttt{General/Specific} label, high-level questions like "How well can it visualize my idea from my words?" are classified as \texttt{General}, while questions focused on specific applications, such as "How well can the model generate game characters with intricate details, like armor or facial expressions?" are labeled as \texttt{Specific Domain}. The \texttt{Specific Domain} label further identifies the focus area for these questions, covering fields like \texttt{Law}, \texttt{Film and Entertainment}, \texttt{Fashion}, \texttt{Game Design}, \texttt{Architecture and Interior Design}, \texttt{Medical}, \texttt{Science and Education}, and \texttt{History and Culture}. We visualize the statistical distribution of the dataset across these categories in Figure~\ref{fig:open_domain_stat}.

\section{More Related Work}
\label{appendix d}
\subsection{Agents Planning \& Reasoning Methods}
The Agent is designed to match human intelligence in decision-making and reasoning, leveraging the core capabilities of LLMs. Several design paradigms in agent designed to boost the performance in agentic systems have been explored.
Tree of Thoughts (ToT)~\cite{yao2024tree} advances the process by constructing a tree-like reasoning structure, where each node represents a reasoning thought, and the final plan is derived through either a breadth-first search (BFS) or depth-first search (DFS) strategy. Algorithm of Thoughts (AoT)~\cite{sel2023algorithm} and Graph of Thoughts (GoT)~\cite{besta2024graph} are the descending works which propel LLMs through algorithmic reasoning pathways and expand the tree-like reasoning structure to a graph-like one respectively. Diagram of Thoughts (DoT)~\cite{zhang2024diagram} is a recently proposed approach that models the reasoning process as a directed acyclic graph (DAG) within a single model, effectively reducing circular dependencies and reflecting well-founded logical deduction.

The general idea of the agent is to take the free-form natural language inputs from the users, plan accordingly, and take action, where LLMs are commonly used as the reasoning and planning backbones. 

To enhance the capability of agents in long-chain reasoning tasks, or a generally defined task with compositionality, humans tend to decompose it into simpler subtasks and solve them procedurally, which also triggers the development of a series of works that mimic the reasoning chain from humans.

Chain-of-Thought (CoT)~\cite{wei2022chain} and Zero-shot-CoT~\cite{kojima2022large} both leverage prompting to trigger them reasoning "step by step", while HuggingGPT~\cite{shen2024hugginggpt} decomposes the tasks into sub-tasks first and solves them independently with Huggingface. However, although those methods attempted to mimic the human thinking process by decomposing tasks and solving each independently, they are still connected in a cascading format, producing only a single-path reasoning chain. Self-Consistent CoT (CoT-SC)~\cite{wang2022self} enhances the original CoT approach by generating multiple reasoning paths and selecting the final answer based on majority voting. Given the trade-off between time and performance, we found that the CoT framework is particularly well-suited for evaluation tasks, as its reasoning process aligns closely with the nature of these tasks.

The reasonings are purely defined by the reasoning backbone of the core LLMs, without incorporating the feedback from either environments or agents themselves. Under this setting, agents' reasoning is straightforward but less effective for long-horizon tasks as it requires the agent to generate flawless plan at the initial stage and it cannot tackle the intermediate failures properly. 
ReAct~\cite{yao2022react} proposed a general paradigm for an agent prompting design by integrating the reasoning traces and task-specific actions in an interleaved triplets ``thought-action-observation" to involve the environmental feedback;  SelfCheck~\cite{miao2023selfcheck} allows agents to review and assess their reasoning steps at different stages, allowing them to identify and correct errors by comparing the intermediate results; Reflexion~\cite{shinn2024reflexion} utilizes ``verbal'' reinforcement learning to augment LLMs with memory encodings. By introducing the feedback into the reasoning steps, agents are equipped with additional knowledge from the feedback to do the correction.

\subsection{Agent in Action Modelling with Tool-Use}
An important factor that differentiates the assistant and agent could be the action modeling capability. Agents should inherently possess the ability to perceive from the environment and interact with the environment via proposed actions~\cite{xi2023rise, wang2024survey}. A trending approach to model the action goal of an agent is the tool-use functionality. \cite{li2023api, qin2023toolllm} proposed benchmarks that can be used to evaluate the tool-use capabilities from the perspectives of API calling functions, requiring the agents to generate or select the appropriate API calls for various tasks and domains based on the natural language inputs. A lot of model-based works also highlights the tool-use functionality, Toolformer~\cite{schick2024toolformer} trained a model in a self-supervised manner to enhance the token prediction while maintaining the generality. 
Gorilla~\cite{patil2023gorillalargelanguagemodel} is a finetuned LLaMA-based model that surpasses the performance of GPT-4 on writing API calls.

\subsection{Recent Works on LLM-based Evaluation for Visual Generation}
Recent works explore LLM-based agents for evaluating visual generative models. \cite{qin2024evaluating} propose a question-answering agent that uses scene graphs to detect hallucinations in text-to-image diffusion models. \cite{yang2025videogen} introduce VideoGen-Eval, a benchmark for text-to-video generation that evaluates different aspects of video quality using LLM-guided reasoning. These studies reflect a growing interest in agent-based, interpretable evaluation frameworks for visual generation tasks.

\section{More Results}
\label{appendix e}

\begin{figure*}
    \centering
    \includegraphics[width=\linewidth]{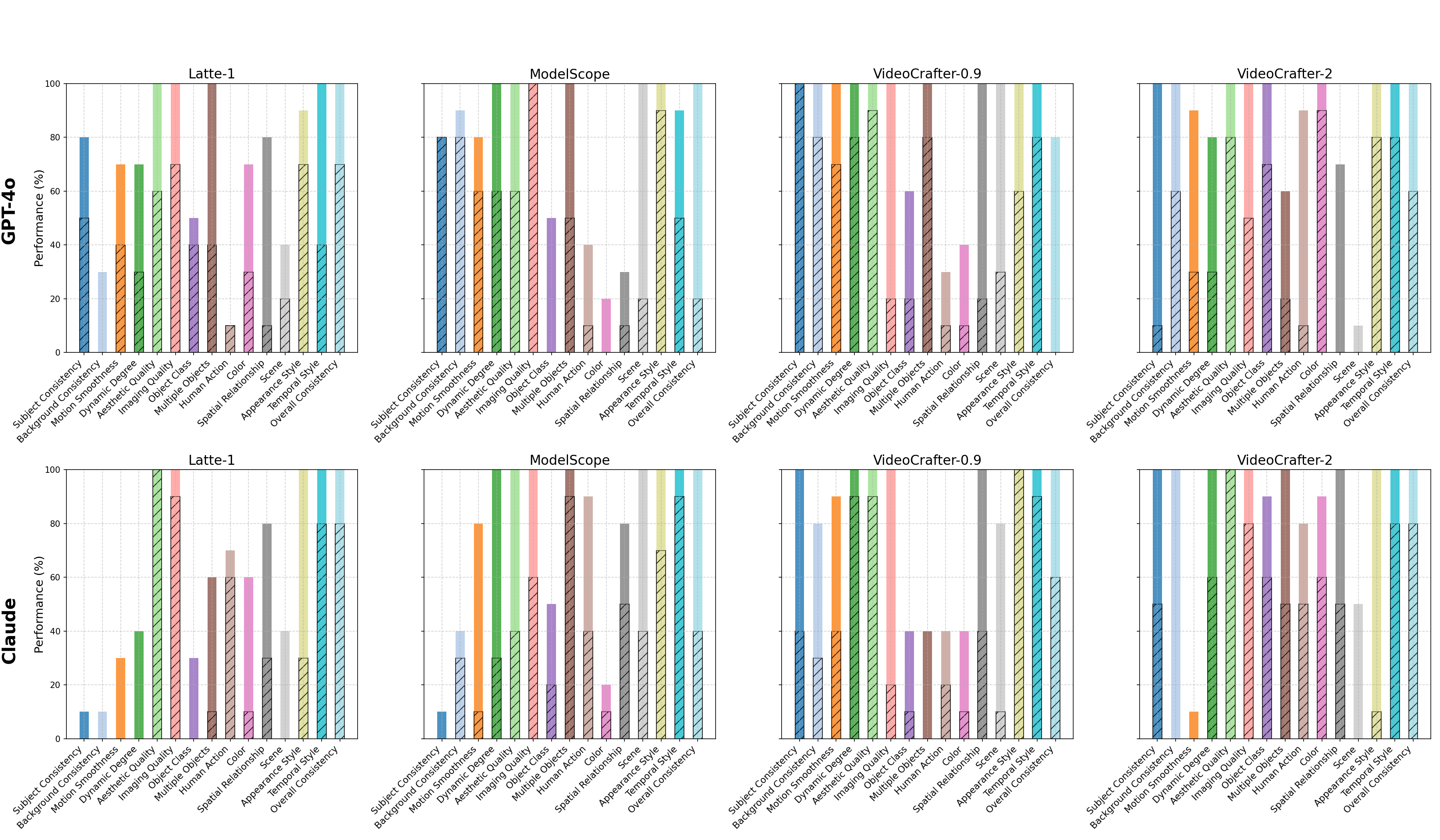}
    \caption{\textbf{Performance Comparison across VBench Dimensions for Different Base Models.} This visualization highlights the performance of all backbone models, including GPT-4o and Claude models, providing a comprehensive comparison in each dimension for different backbone models. Hatched portions indicate predictions within the exact range, and solid portions within an error margin of one range. Specific numerical results are provided in Table~\ref{tab:abl_results_t2v_claude} and Table~\ref{tab:abl_results_t2i_claude}}
    \label{fig:compare_two_base}
\end{figure*}

\subsection{Experiments on Different Base Models}
\label{appendix e1}
We conducted additional experiments using various base models, including API-based models such as Claude and Gemini, to demonstrate the high extensibility of our framework.

\noindent\textbf{Claude.} We conducted validation experiments using the same setup as described in Section~\ref{ssec:closed-domain}, but replaced the base model from \texttt{GPT-4o} to \texttt{Claude-3.5-Sonnet}, specifically using the \texttt{claude-3-5-sonnet-20241022} version. Table~\ref{tab:t2v_time_claude} and Table~\ref{tab:t2i_time_claude} respectively present the time cost and the number of samples required for evaluations across VBench~\cite{huang2023vbench} and T2I-CompBench~\cite{huang2023t2icompbench} dimensions when the base model is replaced with Claude. Furthermore, Table ~\ref{tab:abl_results_t2v_claude} and Table~\ref{tab:abl_results_t2i_claude} provide the evaluation accuracy results corresponding to these experiments. Notably, even with Claude as the base model, only a small number of samples and a few minutes are needed to achieve high-quality evaluation results. Figure~\ref{fig:compare_two_base} illustrates the evaluation outcomes using bar charts, comparing Claude and GPT as the base model.

\begin{figure}[H]
    \centering
    \includegraphics[width=\linewidth]{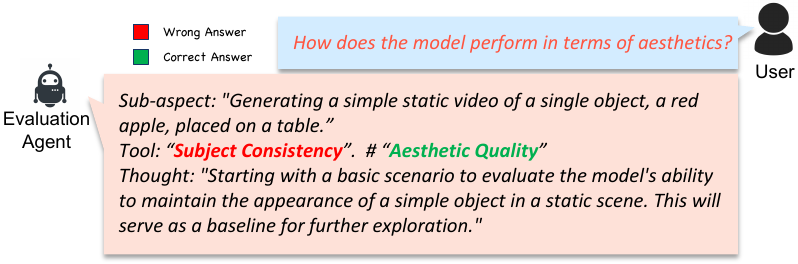}
    \caption{\textbf{A Common Failure Pattern in Tool Selection.} As shown in the figure, Gemini frequently selected an incorrect tool for evaluation. In this case, the model should have selected the “Aesthetic Quality” tool, but it incorrectly chose “Subject Consistency,” leading to inaccuracies in subsequent assessments.}
    \label{fig:gemini_fail1}
\end{figure}

\begin{figure}[H]
    \centering
    \includegraphics[width=\linewidth]{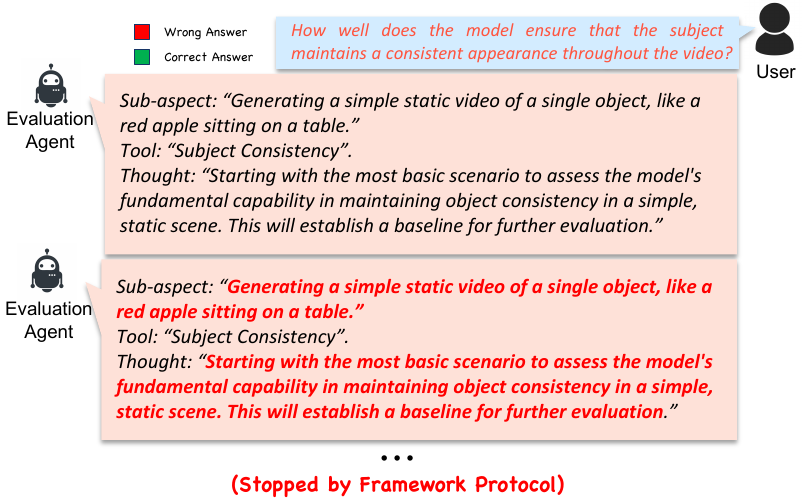}
    \caption{\textbf{Common Failures in Generating Sub-Aspects and Finalizing Responses.} The figure highlights two critical failures: first, Gemini fails to propose new sub-aspects based on observations from previous rounds, instead engaging in repetitive and meaningless loops without strictly adhering to the provided instructions. Second, this repetitive behavior leads to a non-stopping loop, ultimately failing to generate a meaningful final response to the user’s query.}
    \label{fig:gemini_fail2}
\end{figure}

\noindent\textbf{Gemini.} In addition to using Claude, we also experimented with the current state-of-the-art model from the Gemini family, specifically \texttt{gemini-1.5-pro}. However, we found that the evaluation results using Gemini were not ideal, as Gemini often failed to conduct complete and meaningful assessments. Through observation and analysis of the encountered issues, we identified the following common problems: \textbf{\textit{1) Incorrect Tool Selection}:} During the evaluation process, Gemini frequently selected the wrong tool for assessment, despite our detailed and accurate descriptions of each tool provided in the system prompt. For example, as shown in Figure~\ref{fig:gemini_fail1}, when asked, “How does the model perform in terms of aesthetics?”, the evaluation should have used the tool corresponding to the “Aesthetic Quality” dimension. However, Gemini incorrectly chose the tool for the “Subject Consistency” dimension. \textbf{\textit{2) Repetitive and Ineffective Loops}:} We observed that Gemini often generated repetitive or nearly identical proposals (sub-aspects and corresponding thoughts for each round) without adjusting the evaluation direction based on intermediate results. This led to meaningless assessments. For instance, the example in Figure~\ref{fig:gemini_fail2} demonstrates that the sub-aspect and thought proposed in the second round were identical to those in the first round. \textbf{\textit{3) Failure to Generate Final Results}:} The repetitive behavior mentioned in the second issue frequently resulted in a non-stopping loop, ultimately failing to generate a meaningful final response to the user’s query. These issues significantly hinder Gemini’s ability to produce meaningful evaluation results when used as the base model.

We try to analyze the failure modes here for the failure patterns. Although the Gemini model, designed with a context window capable of handling up to 2 million tokens, certain input prompts might still be inadequately comprehended, particularly in complex evaluation tasks. Theoretically, Gemini's extensive context window allows it to process lengthy inputs effectively. However, empirical evidence suggests that even models with large context windows may experience performance degradation when presented with extremely long prompts. This phenomenon is not solely attributable to the absolute length of the input but is also linked to the distribution of critical information across the context. Most transformer-based models, including Gemini, rely on positional encodings to maintain the sequential order of tokens. These encodings inherently influence how the model prioritizes and retrieves information. When key information is placed at the beginning of a prompt, it benefits from being processed early in the attention mechanism. Conversely, information buried deep within a long input might receive attenuated attention, despite the model's theoretical capacity to consider the entire input. In evaluation agent tasks, the agent perceives the instructions with all rules and references set at the beginning by the system prompt, which means some of the essential details are positioned at the beginning of the prompt. Gemini might deprioritize them during the attention allocation process, which could lead to incomplete comprehension, affecting the quality of the evaluation.

In summary, given these findings, we recommend using GPT-4o as the base model for tasks requiring high precision and alignment, particularly for generating prompts that effectively sample from generative models. Its superior performance in producing accurate, contextually aligned prompts contributes to better overall results. Nevertheless, the flexibility of our framework allows for seamless integration with other advanced models like Claude, ensuring its applicability across diverse scenarios and system configurations. This adaptability makes it a valuable tool in environments where model availability or performance priorities may vary.

\subsection{Agent Reliability and Error Handling} 
While the Plan Agent is designed to reason in a structured, multi-step manner, agentic reasoning processes can still encounter failure modes, such as making unreasonable decisions, entering repetitive loops, or failing to converge on a clear evaluation. To mitigate these risks, our framework integrates several safeguards at both the agent and system levels. At the agent level, we adopt a ReAct-style prompting strategy, which encourages the Plan Agent to reason step-by-step and to explicitly articulate its thought process before each action. In particular, the agent is required to provide a justification (“thought”) before terminating the evaluation, helping ensure that such decisions are based on collected evidence rather than arbitrary heuristics. At the system level, we implement hard constraints such as a maximum number of evaluation rounds to prevent infinite loops and control evaluation cost. This has shown practical effectiveness, for example, as illustrated in Figure~\ref{fig:gemini_fail2}, where a repetitive loop triggered by Gemini was successfully halted by the framework protocol. Looking ahead, we believe it is valuable to further explore safeguard strategies for agent-based evaluation. While our current mechanisms offer basic robustness, improving fallback design and reliability remains an important direction for future work, especially as agent-based reasoning becomes more widely adopted.

\subsection{Evaluation Results of Open-Ended User Queries} 
\label{open-results}
\begin{figure*}[htbp!]
    \centering
    \includegraphics[width=0.77\linewidth]{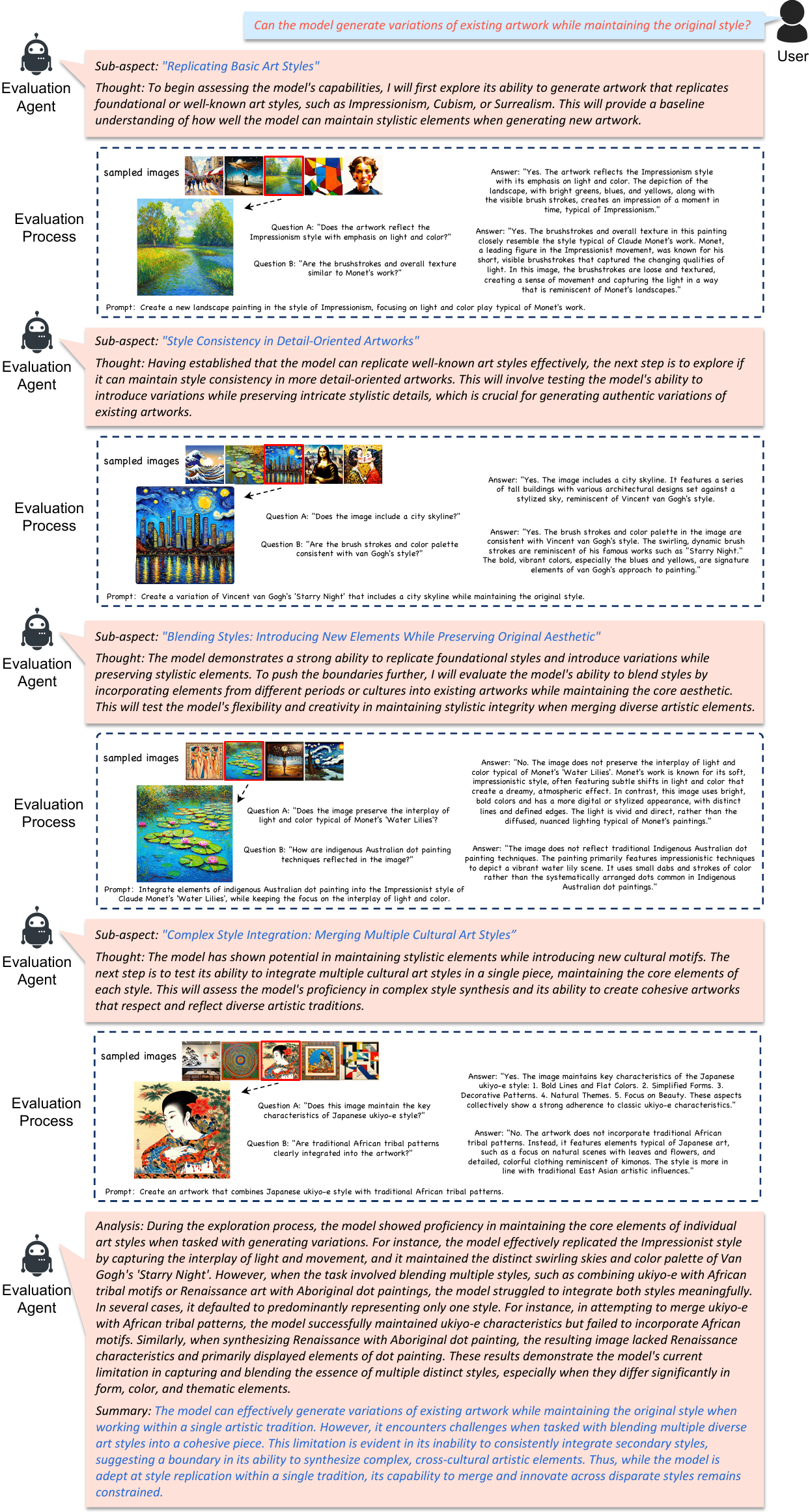}
    \caption{\textbf{A Case of Open-Ended User Query Evaluation.} This figure illustrates the Evaluation Agent’s response to the user query, “Can the model generate variations of existing artwork while maintaining the original style?”}
    \label{fig:appendix_open_case_1}
    \vspace{-20pt}
\end{figure*}

\begin{figure*}[htbp!]
    \centering
    \includegraphics[width=0.77\linewidth]{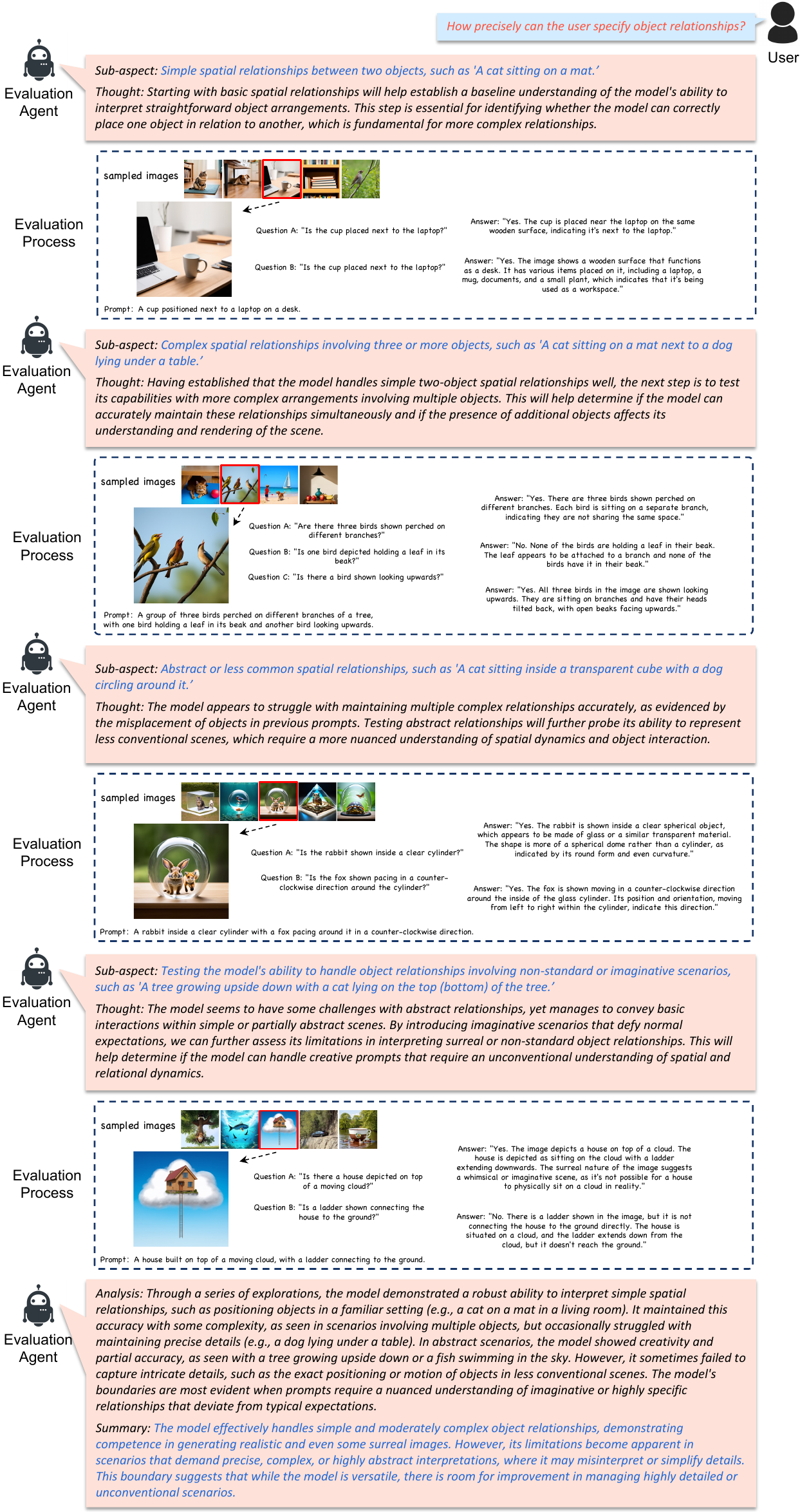}
    \caption{\textbf{A Case of Open-Ended User Query Evaluation.} This figure illustrates the Evaluation Agent’s response to the user query, “How precisely can the user specify object relationships?”}
    \label{fig:appendix_open_case_2}
    \vspace{-20pt}
\end{figure*}

\begin{figure*}[htbp!]
    \centering
    \includegraphics[width=0.77\linewidth]{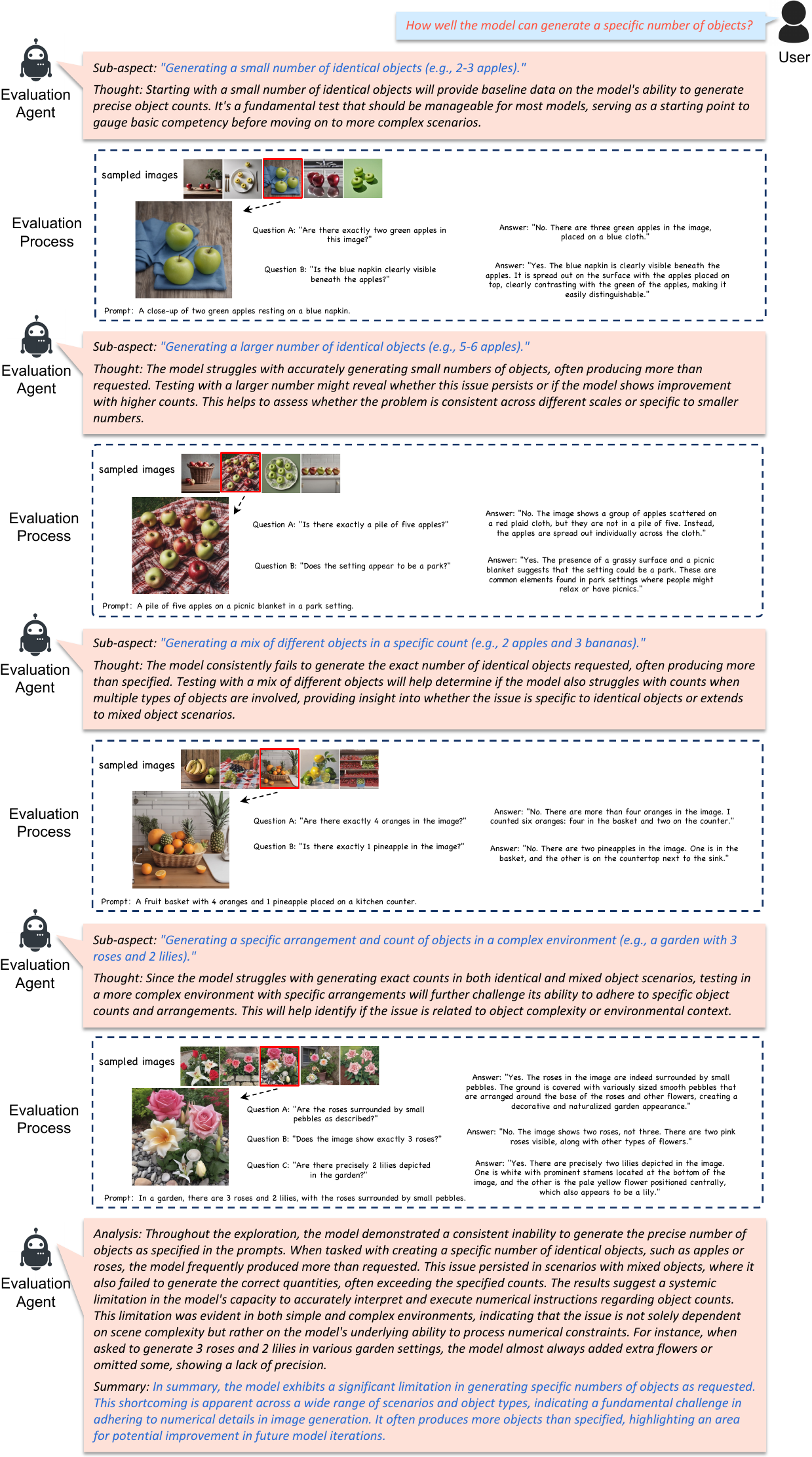}
    \caption{\textbf{A Case of Open-Ended User Query Evaluation.} This figure illustrates the Evaluation Agent’s response to the user query, “How well the model can generate a specific number of objects?”}
    \label{fig:appendix_open_case_3}
    \vspace{-20pt}
\end{figure*}

In Figures~\ref{fig:appendix_open_case_1}-\ref{fig:appendix_open_case_3}, we present multiple comprehensive evaluation results for open-ended user queries, including the sub-aspects raised by the Evaluation Agent in each evaluation round, the corresponding thoughts, the step-by-step evaluation process for each round, and the final comprehensive conclusion.

\end{document}